\documentclass[]{article}
\usepackage{proceed2e}
\usepackage{times}
\usepackage{natbib}
\usepackage[ruled,vlined]{algorithm2e}
\usepackage{amsmath}
\usepackage{amssymb}
\usepackage{graphicx}
\usepackage{wrapfig}

\title{Decentralized Data Fusion and Active Sensing with Mobile Sensors for Modeling and Predicting Spatiotemporal Traffic Phenomena} 

\author{\hspace{-12mm}{\bf Jie Chen$^{\dag}$, Kian Hsiang Low$^{\dag}$, Colin Keng-Yan Tan$^{\dag}$,}\\
\hspace{-12mm}{\bf Ali Oran$^{\S}$}\\
       \hspace{-12mm}Dept. Computer Science, National University of Singapore$^\dag$\\
       \hspace{-12mm}Singapore-MIT Alliance for Research and Technology$^{\S}$\\
       \hspace{-12mm}Republic of Singapore
\And \hspace{12mm}
{\bf Patrick Jaillet$^{\ddag}$, John Dolan$^{\P}$, Gaurav Sukhatme$^{\ast}$}\\  
       \hspace{12mm}Dept. Electrical Engineering and Computer Science, MIT$^{\ddag}$\\
       \hspace{12mm}The Robotics Institute, Carnegie Mellon University$^{\P}$\\
       \hspace{12mm}Dept. Computer Science, University of Southern California$^{\ast}$\\
       \hspace{12mm}USA            
} 

\newcounter{sol} 
\begin{document} 
\newcommand{\set}[1]{{\mathcal #1}}
\newcommand{\squishlisttwo}{
 \begin{list}{$\bullet$}
  { \setlength{\itemsep}{0pt}
     \setlength{\parsep}{0pt}
    \setlength{\topsep}{0pt}
    \setlength{\partopsep}{0pt}
    \setlength{\leftmargin}{1em}
    \setlength{\labelwidth}{1.5em}
    \setlength{\labelsep}{0.5em} } }

\newcommand{\squishend}{
  \end{list}  }

\newtheorem{defi}{Definition}
\newtheorem{thm}{Theorem}
 
\maketitle 
 
\begin{abstract}
\vspace{-5mm}
The problem of modeling and predicting spatiotemporal traffic phenomena over an urban road network is important to many traffic applications such as detecting and forecasting congestion hotspots. This paper presents a decentralized data fusion and active sensing (D$^2$FAS) algorithm for mobile sensors to actively explore the road network to gather and assimilate the most informative data for predicting the traffic phenomenon. 
We analyze the time and communication complexity of D$^2$FAS and demonstrate that it can scale well with a large number of observations and sensors.
We provide a theoretical guarantee on its predictive performance to be equivalent to that of a sophisticated centralized sparse approximation for the Gaussian process (GP) model: The computation of such a sparse approximate GP model can thus be parallelized and distributed among the mobile sensors (in a Google-like MapReduce paradigm), thereby achieving efficient and scalable prediction.
We also theoretically guarantee its active sensing performance that improves under various practical environmental conditions.
Empirical evaluation on real-world urban road network data shows that our D$^2$FAS algorithm
is significantly more time-efficient and scalable than state-of-the-art centralized algorithms while achieving comparable predictive performance.\vspace{-3mm}
\end{abstract}

\section{Introduction}
\label{sec:intro}
\vspace{-3mm}
Knowing and understanding the traffic conditions and phenomena over road networks has become increasingly important to the goal of achieving smooth-flowing, congestion-free traffic, especially in densely-populated urban cities.
According to a $2011$ urban mobility report \citep{Schrank11}, traffic congestion in the USA has caused $1.9$ billion gallons of extra fuel, $4.8$ billion hours of travel delay, and $\$101$ billion of delay and fuel cost. 
Such huge resource wastage can potentially be mitigated if the spatiotemporally varying traffic phenomena (e.g., speeds and travel times along road segments) 
are predicted accurately enough in real time to detect and forecast the congestion hotspots; network-level (e.g., ramp metering, road pricing) and user-level (e.g., route replanning) measures can then be taken to relieve this congestion, so as to improve the overall efficiency of road networks.

In practice, it is non-trivial to achieve real-time, accurate prediction of a spatiotemporally varying traffic phenomenon because the quantity of sensors that can be deployed to observe an entire road network is cost-constrained.
Traditionally, static sensors such as loop detectors \citep{krause08,wang05} are placed at designated locations in a road network to collect data for 
predicting the traffic phenomenon.
However, they provide sparse coverage (i.e., many road segments are not observed, thus leading to data sparsity), incur high installation and maintenance costs, and cannot reposition by themselves in response to changes in the traffic phenomenon.
Low-cost GPS technology allows the collection of traffic data using passive mobile probes \citep{Bayen10a} (e.g., taxis/cabs). 
Unlike static sensors, they can directly measure the travel times along road segments. But, they provide fairly sparse coverage due to low GPS sampling frequency (i.e., often imposed by taxi/cab companies) and no control over their routes, incur high initial implementation cost, pose privacy issues, and produce highly-varying speeds and travel times while traversing the same road segment due to inconsistent driving behaviors. A critical mass of probes is needed on each road segment to ease the severity of the last drawback \citep{Srinivasan96} but is often hard to achieve on non-highway segments due to sparse coverage.
In contrast, we propose the use of active mobile probes \citep{turner98} to overcome the limitations of static and passive mobile probes. In particular, they can be directed to explore any segments of a road network to gather traffic data at a desired GPS sampling rate while enforcing consistent driving behavior.

How then do the mobile probes/sensors actively explore a road network to gather and assimilate the most informative observations for predicting the traffic phenomenon? There are three key issues surrounding this problem, which will be discussed together with the related works:

{\bf Models for predicting spatiotemporal traffic phenomena.}
The spatiotemporal correlation structure of a traffic phenomenon can be exploited to predict the traffic conditions of any unobserved road segment at any time using the observations taken along the sensors' paths. To achieve this, existing Bayesian filtering frameworks \citep{chen11,wang05,Bayen10a}
utilize various handcrafted parametric models predicting traffic flow along a highway stretch
that only correlate adjacent segments of the highway.
So, their predictive performance will be compromised when the current observations are sparse and/or the actual spatial correlation spans multiple segments.
Their strong Markov assumption further exacerbates this problem.
It is also not shown how these models can be generalized to work for arbitrary road network topologies and more complex correlation structure.
Existing multivariate parametric traffic prediction models \citep{kamarianakis03,min11} do not quantify uncertainty estimates of the predictions and impose rigid spatial locality assumptions that do not adapt to the true underlying correlation structure.

In contrast, we assume the traffic phenomenon over an urban road network (i.e., comprising the full range of road types like highways, arterials, slip roads) to be realized from a rich class of Bayesian non-parametric models called the \emph{Gaussian process} (GP) (Section~\ref{sec:gpgk}) that can formally characterize its spatiotemporal correlation structure and be refined with growing number of observations.
More importantly, GP can provide formal measures of predictive uncertainty (e.g., based on variance or entropy criterion) for directing the sensors to explore highly uncertain areas of the road network. 
\cite{krause08} used GP to represent the traffic phenomenon over a network of only highways and defined the correlation of speeds between highway segments to depend only on the geodesic (i.e., shortest path) distance of these segments with respect to the network topology; their features are not considered.
\cite{Kersting09} maintained a mixture of two independent GPs for flow prediction such that the correlation structure of one GP utilized road segment features while that of the other GP depended on manually specified relations (instead of geodesic distance) between segments with respect to an undirected network topology.
Different from the above works, we propose a relational GP
whose correlation structure exploits the geodesic distance between segments based on the topology of a directed road network with vertices denoting road segments and edges indicating adjacent segments weighted by dissimilarity of their features, hence tightly integrating the features and relational information.

{\bf Data fusion.} 
The observations are gathered distributedly by each sensor along its path in the road network and have to be assimilated in order to predict the traffic phenomenon. Since a large number of observations are expected to be collected, a centralized approach to GP prediction cannot be performed in real time due to its cubic time complexity.

To resolve this, we propose a decentralized data fusion approach to efficient and scalable approximate GP prediction (Section~\ref{sec:ddf}).
Existing decentralized and distributed Bayesian filtering frameworks for addressing non-traffic related problems \citep{Chung04,Coates04,saber05a,Thrun03,Durrant-Whyte03b}
will face the same difficulties as their centralized counterparts described above if applied to predicting traffic phenomena, thus resulting in loss of predictive performance.
Distributed regression algorithms \citep{guestrin04,Guestrin04b} for static sensor networks gain efficiency from spatial locality assumptions, which cannot be exploited by mobile sensors whose paths are not constrained by locality.
\cite{cortes09} proposed a distributed data fusion approach to approximate GP prediction based on an iterative Jacobi overrelaxation algorithm, which incurs some critical limitations: (a) the past observations taken along the sensors' paths are assumed to be uncorrelated, which greatly undermines its predictive performance when they are in fact correlated and/or the current observations are sparse;
(b) when the number of sensors grows large, it converges very slowly; 
(c) it assumes that the range of positive correlation has to be bounded by some factor of the communication range. Our proposed decentralized algorithm does not suffer from these limitations and can be computed exactly with efficient time bounds.

{\bf Active sensing.} 
The sensors have to coordinate to actively gather the most informative observations for minimizing the uncertainty of modeling and predicting the traffic phenomenon. Existing centralized \citep{LowAAMAS08,LowICAPS09,LowAAMAS11} and decentralized \citep{,LowAAMAS12,stranders09} active sensing algorithms scale poorly with a large number of observations and sensors.
We propose a partially decentralized active sensing algorithm that overcomes these issues of scalability (Section~\ref{sec:dadgf}).

This paper presents a novel \emph{Decentralized Data Fusion and Active Sensing} (D$^2$FAS) algorithm (Sections~\ref{sec:ddf} and~\ref{sec:dadgf}) for sampling spatiotemporally varying environmental phenomena with mobile sensors. 
Note that the decentralized data fusion component of D$^2$FAS can also be used for static and passive mobile sensors.
The practical applicability of D$^2$FAS is not restricted to traffic monitoring; it can be used in other environmental sensing applications such as mineral prospecting \citep{LowICRA07}, monitoring of ocean and freshwater phenomena \citep{LowSPIE09,LowAeroconf10,LowIPSN09,LowAAMAS11,LowAAMAS12} (e.g., plankton bloom, anoxic zones), forest ecosystems, pollution (e.g., oil spill), or contamination (e.g., radiation leak).
The specific contributions of this paper include:\vspace{-0.6mm}
\squishlisttwo
\item Analyzing the time and communication overheads of D$^2$FAS (Section~\ref{overhead}): We prove that D$^2$FAS can scale better than existing state-of-the-art centralized algorithms with a large number of observations and sensors;
\item Theoretically guaranteeing the predictive performance of the decentralized data fusion component of D$^2$FAS to be equivalent to that of 
a sophisticated centralized sparse approximation for the GP model (Section~\ref{sec:ddf}): The computation of such a sparse approximate GP model can thus be parallelized and distributed among the mobile sensors (in a Google-like MapReduce paradigm), thereby achieving efficient and scalable prediction;
\item Theoretically guaranteeing the performance of the partially decentralized active sensing component of D$^2$FAS, from which various practical environmental conditions can be established to improve its performance;
\item Developing a relational GP model whose correlation structure can exploit both the road segment features
and road network topology information (Section~\ref{sec:gk});
\item Empirically evaluating the predictive performance, time efficiency, and scalability of the D$^2$FAS algorithm on a real-world traffic phenomenon (i.e., speeds of road segments) dataset over an urban road network (Section~\ref{sec:exp}): D$^2$FAS is more time-efficient and scales significantly better with increasing number of observations and sensors while achieving predictive performance close to that of existing state-of-the-art centralized algorithms. \vspace{-1mm}
\squishend
%
\section{Relational Gaussian Process Regression}
\label{sec:gpgk}
\vspace{-3mm}
%
%
The \emph{Gaussian process} (GP) can be used to model a spatiotemporal traffic phenomenon over a road network as follows: The traffic phenomenon is defined to vary as a realization of a GP.
Let $V$ be a set of road segments representing the domain of the road network such that each road segment $s\in V$ is specified by a $p$-dimensional vector of features and is associated with a realized (random) measurement $z_s$ ($Z_s$) of the traffic condition such as speed if $s$ is observed (unobserved). Let $\{Z_s\}_{s \in V}$ denote a GP, that is, every finite subset of $\{Z_s\}_{s \in V}$ follows a multivariate Gaussian distribution \citep{rasmussen06}. Then, the GP is fully specified by its \emph{prior} mean $\mu_{s} \triangleq \mathbb{E}[Z_s]$ and covariance $\sigma_{s s'} \triangleq \mbox{cov}[Z_s, Z_{s'}]$ for all $s, s' \in V$. 
In particular, we will describe in Section~\ref{sec:gk} how the covariance $\sigma_{s s'}$ for modeling the correlation of measurements between all pairs of segments $s, s' \in V$ can be designed to exploit the road segment features and the road network topology.
%
%
%
%
%

A chief capability of the GP model is that of performing probabilistic regression:
Given a set $D\subset V$ of observed road segments and a column vector $z_D$ of corresponding measurements, the joint distribution of the measurements 
at any set $Y\subseteq V\setminus D$ of unobserved road segments remains Gaussian with the following \emph{posterior} mean vector and covariance matrix\vspace{-0mm}
\begin{equation}
\displaystyle\mu_{Y|D}\triangleq \mu_Y + \Sigma_{YD}\Sigma_{DD}^{-1}( z_D - \mu_D )
\label{pmu}\vspace{-0mm}
\end{equation}
\begin{equation}
\displaystyle\Sigma_{YY|D}\triangleq\Sigma_{YY}-\Sigma_{YD}\Sigma_{DD}^{-1}\Sigma_{DY}
\label{pcov}\vspace{-0mm}
\end{equation}
where ${\mu}_Y$ (${\mu}_D$) is a column vector with mean components $\mu_{s}$ for all $s\in Y$ ($s\in D$),
$\Sigma_{YD}$ ($\Sigma_{DD}$) is a covariance matrix with covariance components $\sigma_{s s'}$ for all $s\in Y, s^\prime\in D$ ($s, s^\prime\in D$), and $\Sigma_{DY}$ is the transpose of $\Sigma_{YD}$. The posterior mean vector $\mu_{Y|D}$ (\ref{pmu}) is used to predict the measurements at any set $Y$ of unobserved road segments.
The posterior covariance matrix $\Sigma_{YY|D}$ (\ref{pcov}), which is independent of the measurements $z_D$,
can be processed in two ways to quantify the uncertainty of these predictions: (a) the trace of $\Sigma_{YY|D}$ yields the sum of posterior variances $\Sigma_{ss|D}$ over all $s\in Y$; (b) the determinant of $\Sigma_{YY|D}$ is used in calculating the Gaussian posterior joint entropy
\vspace{-2.5mm}
\begin{equation}
  \mathbb{H}[Z_Y|Z_D]\triangleq
  \frac{1}{2}\log\hspace{-0.5mm}\left(2\pi e\right)^{\left|Y\right|}\left|\Sigma_{YY|D}\right| \ .
  \label{entropy}\vspace{-2.5mm}
\end{equation}
In contrast to the first measure of uncertainty that assumes conditional independence between measurements in the set $Y$ of unobserved road segments, the entropy-based measure (\ref{entropy}) accounts for their correlation, thereby not overestimating their uncertainty.
Hence, we will focus on using the entropy-based measure of uncertainty in this paper.\vspace{-2mm}
%
\subsection{Graph-Based Kernel}
\label{sec:gk}\vspace{-3mm}
If the observations are noisy (i.e., by assuming additive independent identically distributed Gaussian noise with variance $\sigma^2_n$), then their prior covariance $\sigma_{s s'}$ can be expressed as
$
\sigma_{s s'} = k(s,s') + \sigma^2_n \delta_{s s'}
$
where $\delta_{s s'}$ is a Kronecker delta that is $1$ if $s = s'$ and $0$ otherwise, and $k$ is a kernel function measuring the pairwise ``similarity'' of road segments.
For a traffic phenomenon (e.g., road speeds),
the correlation of measurements between pairs of road segments depends not only on their features (e.g., length, number of lanes, speed limit, direction) but also
the road network topology.  
So, the kernel function is defined to exploit both the features and topology information, which will be described next.\vspace{-1.5mm}  
\begin{defi}[Road Network]
Let the road network be represented as a weighted directed graph $G\triangleq (V,E,m)$ that consists of\vspace{-1mm}
\squishlisttwo
\item a set $V$ of vertices denoting the domain of all possible road segments,
\item a set $E\subseteq V\times V$ of edges where there is an edge $(s,s')$ from $s\in V$ to $s'\in V$ iff the end of segment $s$ connects to the start of segment $s'$ in the road network, and
\item a weight function $m: E\rightarrow \mathbb{R}^{+}$ measuring 
the standardized Manhattan distance \citep{borg05}
$m((s,s'))\triangleq\sum_{i=1}^p \left|[s]_i-[s']_i\right|\hspace{-0.5mm}/r_i$
 of each edge $(s,s')$
where $[s]_i$ ($[s']_i$) is the $i$-th component of the feature vector specifying road segment $s$ ($s'$), and $r_i$ is the range of the $i$-th feature. 
The weight function $m$ serves as a dissimilarity measure between adjacent road segments.\vspace{-1.5mm}
\squishend
\end{defi} 
%
%
The next step is to compute the shortest path distance $d(s,s')$ between all pairs of road segments $s,s'\in V$ (i.e., using Floyd-Warshall or Johnson's algorithm) with respect to the topology of the weighted directed graph $G$.
Such a distance function is again a measure of dissimilarity, rather than one of similarity, as required by a kernel function. 
Furthermore, a valid GP kernel needs to be positive semidefinite and symmetric \citep{scholkopf02}, which are clearly violated by $d$.  

To construct a valid GP kernel from $d$, multi-dimensional scaling \citep{borg05} is
applied to embed the domain of road segments into the $p'$-dimensional Euclidean space $\mathbb{R}^{p'}$.
Specifically, a mapping $g: V\rightarrow\mathbb{R}^{p'}$ is determined by minimizing the squared loss
%
$  g^\ast=\mathop{\arg\min}_g \sum_{s,s'\in V}
  (d(s,s')-\|g(s)-g(s')\|)^2
$.
%
With a small squared loss, the Euclidean distance $\|g^{\ast}(s)-g^{\ast}(s')\|$ between $g^{\ast}(s)$ and $g^{\ast}(s')$ is expected to closely approximate the shortest path distance $d(s,s')$ between any pair of road segments $s$ and $s'$.
After embedding into Euclidean space, a conventional kernel function such as the squared exponential one \citep{rasmussen06} can be used:\vspace{-2.5mm}
\begin{equation*}
  k(s,s')=\sigma_s^2\exp\left({-\frac{1}{2}\sum_{i=1}^{p'}\left(\frac{[g^{\ast}(s)]_i-[g^{\ast}(s')]_i}{\ell_i}\right)^2}\right)
\label{gkernel}\vspace{-2.5mm}
\end{equation*}
where $[g^{\ast}(s)]_i$ ($[g^{\ast}(s')]_i$) is the $i$-th component of the $p'$-dimensional vector $g^{\ast}(s)$ ($g^{\ast}(s')$), and the hyperparameters $\sigma_s$, $\ell_1, \ldots, \ell_{p'}$ are, respectively, signal variance and length-scales that can be learned using maximum likelihood estimation \citep{rasmussen06}. The resulting kernel function $k$\footnote{For spatiotemporal traffic modeling, the kernel function $k$ can be extended to account for the temporal dimension.} is guaranteed to be valid. \vspace{-2mm}

\subsection {Subset of Data Approximation}
\label{sec:sgp}\vspace{-3mm}
Although the GP is an effective predictive model, it faces a practical limitation of cubic time complexity in the number $|D|$ of observations; this can be observed from computing the posterior distribution (i.e., (\ref{pmu}) and (\ref{pcov})), which requires inverting covariance matrix $\Sigma_{DD}$ and incurs $\set{O}(|D|^3)$ time.
If $|D|$ is expected to be large, GP prediction cannot be performed in real time.  
For practical usage, we have to resort to computationally cheaper approximate GP prediction.

A simple method of approximation is to select only a subset $U$ of the entire set $D$ of observed road segments (i.e., $U\subset D$)
to compute the posterior distribution of the measurements at any set $Y\subseteq V\setminus D$ of unobserved road segments.
Such a \emph{subset of data} (SoD) approximation method produces the following predictive Gaussian distribution, which closely resembles that of the full GP model (i.e., by simply replacing $D$ in (\ref{pmu}) and (\ref{pcov}) with $U$):\vspace{-2mm}
%
\begin{equation}
  \mu_{Y|U}=\mu_Y+\Sigma_{YU}\Sigma_{UU}^{-1} (z_U - \mu_U) 
\label{eq:fsod}\vspace{-2mm}
\end{equation}
\begin{equation}
  \Sigma_{YY|U}=\Sigma_{YY}-\Sigma_{YU}\Sigma_{UU}^{-1}\Sigma_{UY}\ .
\label{pcovsod}\vspace{-1mm}
\end{equation}
Notice that the covariance matrix $\Sigma_{UU}$ to be inverted only incurs $\set{O}(|U|^3)$ time, which is independent of $|D|$.  

The predictive performance of SoD approximation is sensitive to the selection of subset $U$.
In practice, random subset selection often yields poor performance.  
This issue can be resolved by actively selecting an informative subset $U$ in an iterative greedy manner:
Firstly, $U$ is initialized to be an empty set. 
Then, all road segments in $D\setminus U$ are scored based on a criterion that can be chosen from, for example, the works of \citep{Guestrin08,lawrence03,seeger03}.
The highest-scored segment is selected for inclusion in $U$ and removed from $D$.
This greedy selection procedure is iterated until $U$ reaches a pre-defined size.  
Among the various criteria introduced earlier, the differential entropy score \citep{lawrence03} is reported to perform well \citep{oh10}; it is a monotonic function of the posterior variance $\Sigma_{ss|U}$ (\ref{pcovsod}), thus resulting in the greedy selection of a segment $s\in D\setminus U$ with the largest variance in each iteration. \vspace{-3mm}
%
\section{Decentralized Data Fusion}
\label{sec:ddf}\vspace{-3mm}
In the previous section, two centralized data fusion approaches to exact (i.e., (\ref{pmu}) and (\ref{pcov})) and approximate (i.e., (\ref{eq:fsod}) and (\ref{pcovsod})) GP prediction are introduced.
In this section, we will discuss the decentralized data fusion component of our D$^2$FAS algorithm, which distributes the computational load among the mobile sensors to achieve efficient and scalable approximate GP prediction.

The intuition of our decentralized data fusion algorithm is as follows: Each of the $K$ mobile sensors constructs a local summary of the observations taken along its own path in the road network and communicates its local summary to every other sensor. Then, it assimilates the local summaries received from the other sensors into a globally consistent summary, which is exploited for predicting the traffic phenomenon as well as active sensing.
This intuition will be formally realized and described in the paragraphs below.

While exploring the road network, each mobile sensor summarizes its local observations taken along its path based on a common support set $U\subset V$ known to all the other sensors. Its local summary is defined as follows:\vspace{-1.5mm}
%
\begin{defi}[Local Summary]
Given a common support set $U\subset V$ known to all $K$ mobile sensors, a set $D_k \subset V$ of observed road segments and a column vector $z_{D_k}$  of corresponding measurements local to mobile sensor $k$, its local summary is defined as a tuple $(\dot{z}_U^k, \dot{\Sigma}_{UU}^k)$ where\vspace{-0.5mm}
\begin{equation}
\dot{z}_{U}^k \triangleq\Sigma_{UD_k} \Sigma^{-1}_{D_k D_k|U} (z_{D_k} - \mu_{D_k})
\label{lsf}\vspace{-0.5mm}
\end{equation}
\begin{equation}
\dot{\Sigma}_{UU}^k \triangleq\Sigma_{UD_k}\Sigma_{D_k D_k|U}^{-1}\Sigma_{D_k U}
\label{lsc}\vspace{-0.5mm}
\end{equation}
such that $\Sigma_{D_k D_k|U}$ is defined in a similar manner to $($\ref{pcovsod}$)$.
\label{def1}\vspace{-1.5mm}
\end{defi}
\emph{Remark}. Unlike SoD (Section~\ref{sec:sgp}), the support set $U$ of road segments does not have to be observed, since the local summary (i.e., (\ref{lsf}) and (\ref{lsc})) is independent of the corresponding measurements $z_U$. 
So, $U$ does not need to be a subset of $D=\bigcup^K_{k=1}D_k$. To select an informative support set $U$ from the set $V$ of all possible segments in the road network, an offline active selection procedure similar to that in the last paragraph of Section~\ref{sec:sgp} can be performed just once prior to observing data to determine $U$. In contrast, SoD has to perform online active selection every time new road segments are being observed.
%

By communicating its local summary to every other sensor, each mobile sensor can then construct a globally consistent summary from the received local summaries:\vspace{-1.5mm}
\begin{defi}[Global Summary] Given a common support set $U\subset V$ known to all $K$ mobile sensors and the local summary $(\dot{z}_U^k, \dot{\Sigma}_{UU}^k)$ of every mobile sensor $k = 1,\ldots,K$,
the global summary 
is defined as a tuple $(\ddot{z}_U, \ddot{\Sigma}_{UU})$ where\vspace{-2mm}
\begin{equation}
\ddot{z}_U \triangleq\sum_{k=1}^{K}\dot{z}_{U}^k
\label{gsf}\vspace{-2mm}
\end{equation}
\begin{equation}
\ddot{\Sigma}_{UU} \triangleq\Sigma_{UU}+\sum_{k=1}^{K} \dot{\Sigma}_{UU}^k \ .
\label{gsc}\vspace{-6mm}
\end{equation}
\label{def2}
\end{defi}
\emph{Remark}. In this paper, we assume all-to-all communication between the $K$ mobile sensors.
Supposing this is not possible and each sensor can only communicate locally with its neighbors, the summation structure of the global summary (specifically, (\ref{gsf}) and (\ref{gsc})) makes it amenable to be constructed using distributed consensus filters \citep{saber05a}. We omit these details since they are beyond the scope of this paper.

Finally, the global summary is exploited by each mobile sensor to compute a globally consistent predictive Gaussian distribution, as detailed in Theorem~\ref{thmdad}A below, as well as to perform decentralized active sensing (Section~\ref{sec:dadgf}):\vspace{-2mm}
\begin{thm}
Let a common support set $U\subset V$ be known to all $K$ mobile sensors.\vspace{-4mm}
\begin{description}
\item[A.\ \ ] Given the global summary $(\ddot{z}_U, \ddot{\Sigma}_{UU})$, each mobile sensor computes a globally consistent predictive Gaussian distribution
  $\mathcal{N}(\overline{\mu}_{Y},\overline{\Sigma}_{YY})$ of the measurements at any set $Y$ of unobserved road segments where\vspace{-1mm}
%
\begin{equation}
    \overline{\mu}_{Y} \triangleq\mu_Y + \Sigma_{YU}\ddot{\Sigma}_{UU}^{-1}\ddot{z}_U
\label{mupitc}\vspace{-1mm}
\end{equation}
\begin{equation}
  \overline{\Sigma}_{YY} \triangleq\Sigma_{YY}-\Sigma_{YU}(\Sigma_{UU}^{-1}-\ddot{\Sigma}
_{UU}^{-1})\Sigma_{UY} \ .
\label{varpitc}\vspace{-0mm}
\end{equation}
\item[B.\ \ ] Let $\mathcal{N}(\mu_{Y|D}^{\mbox{\tiny\emph{PITC}}},\Sigma_{YY|D}^{\mbox{\tiny\emph{PITC}}})$ be  the predictive Gaussian distribution computed by the centralized sparse partially independent training conditional (PITC) approximation of GP model \citep{candela05} where\vspace{-2mm}
\begin{equation} 
  \mu_{Y|D}^{\mbox{\tiny\emph{PITC}}} \triangleq\mu_Y+\Gamma_{YD}\left(\Gamma_{DD}+\Lambda\right)^{-1}\hspace{-1mm}\left(z_D-\mu_D \right)
  \label{eq:pitc.mu}\vspace{-2mm}
\end{equation}   
\begin{equation}   
\Sigma_{YY|D}^{\mbox{\tiny\emph{PITC}}} \triangleq\Sigma_{YY}-\Gamma_{YD}\left(\Gamma_{DD}+\Lambda\right)^{-1}\Gamma_{DY}
  \label{eq:pitc.var} \vspace{-0mm}
\end{equation} 
such that \vspace{-0mm}
\begin{equation} 
  \Gamma_{BB'} \triangleq\Sigma_{BU}\Sigma_{UU}^{-1}\Sigma_{UB'} 
  \label{eq:sparseq}
\end{equation} 
and 
$\Lambda$
is a 
block-diagonal matrix constructed from the $K$ diagonal blocks of $\Sigma_{DD|U}$,
each of which is a 
matrix $\Sigma_{D_k D_k|U}$
for $k=1,\ldots,K$ where $D=\bigcup_{k=1}^K D_k$.
Then, $\overline{\mu}_{Y}\hspace{-0.5mm}=\hspace{-0.5mm}\mu_{Y|D}^{\mbox{\tiny\emph{PITC}}}$ and $\overline{\Sigma}_{YY}\hspace{-0.5mm}=\hspace{-0.5mm}\Sigma_{YY|D}^{\mbox{\tiny\emph{PITC}}}$.
\end{description} 
\label{thmdad}\vspace{-4mm}
\end{thm}
%
\ifthenelse{\value{sol}=1}{The proof of Theorem~\ref{thmdad}B is given in Appendix~\ref{equ}.} 
{Its proof is given in \citep{LowArxiv12}.} 
The equivalence result of Theorem~\ref{thmdad}B bears two implications:

\emph{Remark} $1$.
The computation of PITC can be parallelized and distributed among the mobile sensors in a Google-like MapReduce paradigm \citep{Chu07b}, thereby improving the time efficiency of prediction: Each of the $K$ mappers (sensors) is tasked to compute its local summary while the reducer (any sensor) sums these local summaries into a global summary, which is then used to compute the predictive Gaussian distribution.
Supposing $|Y|\leq |U|$ for simplicity, the $\set{O}\hspace{-1mm}\left(|D|((|D|/K)^2+|U|^2)\right)$ time incurred by PITC can be reduced to $\set{O}\hspace{-1mm}\left((|D|/K)^3+|U|^3+|U|^2 K\right)$ time of running our decentralized algorithm on each of the $K$ sensors, the latter of which scales better with increasing number $|D|$ of observations.

\emph{Remark} $2$.
We can draw insights from PITC to elucidate an underlying property of our decentralized algorithm: 
It is assumed that $Z_{D_1},\ldots,Z_{D_K}, Z_Y$ are conditionally independent given the measurements at the support set $U$ of road segments.
To potentially reduce the degree of violation of this assumption, an informative support set $U$ is actively selected, as described earlier in this section.
Furthermore, the experimental results on real-world urban road network data\footnote{\cite{candela05} only illustrated the predictive performance of PITC on a simulated toy example.} (Section~\ref{sec:exp})
show that D$^2$FAS can achieve predictive performance comparable to that of the full GP model while enjoying significantly lower computational cost, thus demonstrating the practicality of such an assumption for predicting traffic phenomena.
The predictive performance of D$^2$FAS can be improved by increasing the size of $U$ at the expense of greater time and communication overhead. \vspace{-2mm}
%
\section{Decentralized Active Sensing}
\label{sec:dadgf}\vspace{-3mm}
The problem of active sensing with $K$ mobile sensors is formulated as follows: 
Given the set $D_k\subset V$ of observed road segments and the currently traversed road segment $s_k\in V$ of every mobile sensor $k =1,\ldots,K$, the mobile sensors have to coordinate to select the most informative walks $w^{\ast}_1,\ldots, w^{\ast}_K$ of length (i.e., number of road segments) $L$ each and with respective origins $s_1,\ldots,s_K$ in the road network $G$: \vspace{-1mm}
%
\begin{equation}
\hspace{-0.3mm}
\displaystyle(w^{\ast}_1,\ldots, w^{\ast}_K)\hspace{-0.8mm} = \hspace{-0.8mm}\mathop{\arg\max}_{(w_1,\ldots, w_K)}\mathbb{H}\hspace{-1mm}\left[Z_{\bigcup^K_{k=1} Y_{w_k}} \Big{|} Z_{\bigcup^K_{k=1} D_k}\right]\vspace{-1mm}
\label{maxent}
\end{equation}
where $Y_{w_k}$ denotes the set of unobserved road segments induced by the walk $w_k$.
To simplify notation, let a joint walk be denoted by $w\triangleq (w_1,\ldots,w_K)$ (similarly, for $w^\ast$) and its induced set of unobserved road segments be $Y_{w}\triangleq \bigcup^K_{k=1} Y_{w_k}$ from now on.
Interestingly, it can be shown using the chain rule for entropy that these maximum-entropy walks $w^{\ast}$ minimize the posterior joint entropy (i.e., $\mathbb{H}[Z_{V\setminus(D\bigcup Y_{w^{\ast}})} | Z_{D\bigcup Y_{w^{\ast}}}]$) of the measurements at the remaining unobserved segments (i.e., $V\setminus(D\bigcup Y_{w^{\ast}})$) in the road network.
After executing the walk $w^{\ast}_k$, each mobile sensor $k$ observes the set $Y_{w^{\ast}_k}$ of road segments and updates its local information:\vspace{-0.8mm}
%
\begin{equation}
  D_k\leftarrow D_k\bigcup Y_{w^{\ast}_k} \ , z_{D_k}\leftarrow z_{D_k\bigcup Y_{w^{\ast}_k}}, s_k\leftarrow \mbox{terminus of}\ w^{\ast}_k\ .\vspace{-0mm}
\label{upDk}
\end{equation}
%
%
Evaluating the Gaussian posterior entropy term in (\ref{maxent}) involves computing a posterior covariance matrix (\ref{entropy}) using one of the data fusion methods described earlier:
If (\ref{pcov}) of full GP model (Section~\ref{sec:gpgk}) or (\ref{pcovsod}) of SoD (Section~\ref{sec:sgp}) is to be used, then the observations that are gathered distributedly by the sensors have to be fully communicated to a central data fusion center.
In contrast, our decentralized data fusion algorithm (Section~\ref{sec:ddf}) only requires communicating local summaries (Definition~\ref{def1}) to compute (\ref{varpitc}) for solving the active sensing problem (\ref{maxent}):\vspace{-0.5mm}
\begin{equation}
  \displaystyle w^{\ast}=\displaystyle\mathop{\arg\max}_{w}\ \overline{\mathbb{H}}\hspace{-0.5mm}\left[Z_{Y_w}\right]\ ,\label{maxent2}\vspace{-1.5mm}
\end{equation}
\begin{equation}
\overline{\mathbb{H}}\hspace{-0.5mm}\left[Z_{Y_w}\right]\triangleq\frac{1}{2}\log\hspace{-0.5mm}\left(2\pi e\right)^{\left|Y_w\right|}\left|\overline{\Sigma}_{Y_{w} Y_{w}}\right|\ .\label{maxent21}\vspace{-1mm}
\end{equation}
Without imposing any structural assumption, solving the active sensing problem (\ref{maxent2}) will be  prohibitively expensive due to the space of possible joint walks $w$ that grows exponentially in the number $K$ of mobile sensors.
To overcome this scalability issue for D$^2$FAS, our key idea is to construct a block-diagonal matrix whose log-determinant closely approximates that of $\overline{\Sigma}_{Y_{w} Y_{w}}$ (\ref{varpitc}) and exploit the property that the log-determinant of such a block-diagonal matrix can be decomposed into a sum of log-determinants of its diagonal blocks, each of which depends only on the walks of a disjoint subset of the $K$ mobile sensors.
Consequently, the active sensing problem can be partially decentralized, leading to a reduced space of possible joint walks to be searched, as detailed in the rest of this section. 

Firstly, we extend an earlier assumption in Section~\ref{sec:ddf}:
$Z_{D_1},\ldots,Z_{D_K}, Z_{Y_{w_1}},\ldots,Z_{Y_{w_K}}$ are conditionally independent given the measurements at the support set $U$ of road segments. 
Then, it can be shown via the equivalence to PITC (Theorem~\ref{thmdad}B) 
that $\overline{\Sigma}_{Y_{w} Y_{w}}$ (\ref{varpitc}) comprises diagonal blocks of the form $\overline{\Sigma}_{Y_{w_k} Y_{w_k}}$ for $k=1,\ldots,K$ and off-diagonal blocks of the form $\Sigma_{Y_{w_k}U}\ddot{\Sigma}
_{UU}^{-1}\Sigma_{UY_{w_{k'}}}$ for $k,k'=1,\ldots,K$ and $k\neq k'$.
In particular, each off-diagonal block of $\overline{\Sigma}_{Y_{w} Y_{w}}$ represents the correlation of measurements between the unobserved road segments $Y_{w_k}$ and $Y_{w_{k'}}$ along the respective walks $w_k$ of sensor $k$ and $w_{k'}$ of sensor $k'$. If the correlation between some pair of their possible walks is high enough, then
their walks have to be coordinated.
This is formally realized by the following coordination graph over the $K$ sensors:\vspace{-1.5mm}
\begin{defi}[Coordination Graph]\label{coord}
Define the coordination graph to be an undirected graph $\mathcal{G}\triangleq (\mathcal{V}, \mathcal{E})$ that comprises\vspace{-2mm}
\squishlisttwo
\item a set $\mathcal{V}$ of vertices denoting the $K$ mobile sensors, and 
\item a set $\mathcal{E}$ of edges denoting coordination dependencies between sensors such that there exists an edge $\{k,k'\}$ incident with sensors $k\in \mathcal{V}$ and $k'\in \mathcal{V}\setminus\{k\}$ iff\vspace{-1mm}
\begin{equation}
\max_{s\in Y_{W_k}, s'\in Y_{W_{k'}}} \left|\Sigma_{sU}\ddot{\Sigma}_{UU}^{-1}\Sigma_{Us'}\right| > \varepsilon
\label{rel}\vspace{-1mm}
\end{equation}
for a predefined constant $\varepsilon >0$ where $W_k$ denotes the set of possible walks of length $L$ of mobile sensor $k$ from origin $s_k$ in the road network $G$ and $Y_{W_k} \triangleq \bigcup_{w_k\in W_k}Y_{w_k}$.\vspace{-1mm}
\squishend
\end{defi}
\emph{Remark}. The construction of $\mathcal{G}$ can be decentralized as follows:
Since $\ddot{\Sigma}_{UU}$ is symmetric and positive definite,
it can be decomposed by Cholesky factorization into $\ddot{\Sigma}_{UU} = \Psi \Psi^{\top}$ where $\Psi$ is a lower triangular matrix and $\Psi^{\top}$ is the transpose of $\Psi$.
Then,
$\Sigma_{sU}\ddot{\Sigma}_{UU}^{-1}\Sigma_{Us'} =  (\Psi\backslash\Sigma_{Us})^{\top}\Psi\backslash\Sigma_{Us'}$ where $\Psi\backslash B$ denotes the column vector $\phi$ solving $\Psi\phi = B$.
That is, $\Sigma_{sU}\ddot{\Sigma}_{UU}^{-1}\Sigma_{Us'}$ (\ref{rel}) can be expressed as a dot product of two vectors $\Psi\backslash\Sigma_{Us}$ and $\Psi\backslash\Sigma_{Us'}$; this property is exploited to determine adjacency between sensors in a decentralized manner:\vspace{-2mm} 
\begin{defi}[Adjacency]\label{adj}
Let\vspace{-1.5mm}
\begin{equation}
\Phi_k \triangleq\{\Psi\backslash\Sigma_{Us}\}_{s\in Y_{W_k}} 
\label{phik}\vspace{-1.5mm}
\end{equation}
for $k=1,\ldots,K$.
A sensor $k\in \mathcal{V}$ is adjacent to sensor $k'\in \mathcal{V}\setminus\{k\}$ in coordination graph $\mathcal{G}$ iff\vspace{-1.5mm}
\begin{equation}
\max_{\phi\in \Phi_k, \phi'\in \Phi_{k'}} \left|\phi^{\top} \phi'\right| > \varepsilon \ .
\label{rel2}\vspace{-3.0mm}
\end{equation}
\end{defi}
It follows from the above definition that if each sensor $k$ constructs $\Phi_k$ and exchanges it with every other sensor, then it can determine its adjacency to all the other sensors and store this information in a column vector $a_k$ of length $K$ with its $k'$-th component being defined as follows:\vspace{-2mm}
\begin{equation}
\hspace{-1mm}
\left[a_k\right]_{k'}\hspace{-0.1mm} =\hspace{-0.1mm}\Bigg{\{}
\begin{array}{rl}
1 & \mbox{if sensor $k$ is adjacent to sensor $k'$,}\vspace{1mm}\\
0 & \mbox{otherwise.}
\end{array}
\label{adjv}\vspace{-2mm}
\end{equation}
By exchanging its adjacency vector $a_k$ with every other sensor, each sensor can construct a globally consistent adjacency matrix $A_{\mathcal{G}} \triangleq (a_1\ldots a_K)$ to represent coordination graph $\mathcal{G}$.

Next, by computing the connected components (say, $\mathcal{K}$ of them) of coordination graph $\mathcal{G}$, their resulting vertex sets partition the set $\mathcal{V}$ of $K$ sensors into $\mathcal{K}$ disjoint subsets $\mathcal{V}_1,\ldots,\mathcal{V}_{\mathcal{K}}$ such that the sensors within each subset have to coordinate their walks.
Each sensor can determine its residing connected component in a decentralized way by performing a depth-first search in $\mathcal{G}$ starting from it as root.

Finally, construct a block-diagonal matrix $\widehat{\Sigma}_{Y_{w} Y_{w}}$ to comprise diagonal blocks of the form $\overline{\Sigma}_{Y_{w_{\mathcal{V}_n}} Y_{w_{\mathcal{V}_n}}}$ for $n = 1,\ldots,\mathcal{K}$
where $w_{\mathcal{V}_n}\triangleq (w_k)_{k\in \mathcal{V}_n}$ and $Y_{w_{\mathcal{V}_n}}\triangleq\bigcup_{k\in \mathcal{V}_n} Y_{w_k}$.
The active sensing problem (\ref{maxent2}) is then approximated by\vspace{-2mm}
\begin{equation}
\hspace{-0.5mm}
\begin{array}{l}
\displaystyle\max_{w}\ \frac{1}{2}\log\hspace{-0.5mm}\left(2\pi e\right)^{\left|Y_w\right|}\left|\widehat{\Sigma}_{Y_{w} Y_{w}}\right|\\ 
\equiv\displaystyle\max_{(w_{\mathcal{V}_1},\ldots, w_{\mathcal{V}_{\mathcal{K}}})}\sum_{n=1}^{\mathcal{K}}\log\hspace{-0.5mm}\left(2\pi e\right)^{\left|Y_{w_{\mathcal{V}_n}}\right|}\left|\overline{\Sigma}_{Y_{w_{\mathcal{V}_n}} Y_{w_{\mathcal{V}_n}}}\right|\\
=\displaystyle\sum_{n=1}^{\mathcal{K}}\max_{w_{\mathcal{V}_n}}\ \log\hspace{-0.5mm}\left(2\pi e\right)^{\left|Y_{w_{\mathcal{V}_n}}\right|}\left|\overline{\Sigma}_{Y_{w_{\mathcal{V}_n}} Y_{w_{\mathcal{V}_n}}}\right|,
\end{array}
\label{bdiaga}\vspace{-2mm}
\end{equation}
which can be solved in a partially decentralized manner by each disjoint subset $\mathcal{V}_n$ of mobile sensors:\vspace{-2mm}
%
\begin{equation}
\displaystyle \widehat{w}_{\mathcal{V}_n} = 
\mathop{\arg\max}_{w_{\mathcal{V}_n}}\ \log\hspace{-0.5mm}\left(2\pi e\right)^{\left|Y_{w_{\mathcal{V}_n}}\right|}\left|\overline{\Sigma}_{Y_{w_{\mathcal{V}_n}} Y_{w_{\mathcal{V}_n}}}\right|.
  \label{dmaxent}\vspace{-2mm}
\end{equation}
Our active sensing algorithm becomes fully decentralized if $\varepsilon$ is set to be sufficiently large: more sensors become isolated in $\mathcal{G}$, consequently decreasing the size $\displaystyle\kappa\triangleq\max_{n}\left|\mathcal{V}_n\right|$ of its largest connected component to $1$.
As shown in Section~\ref{sec:time}, 
decreasing $\kappa$ improves its time efficiency.
On the other hand, it tends to a centralized behavior (\ref{maxent2}) by setting $\varepsilon\rightarrow 0^{+}$: $\mathcal{G}$ becomes near-complete, thus resulting in $\kappa\rightarrow K$. \vspace{-2mm}
%
%

Let
\vspace{-3.5mm} 
\begin{equation}
\displaystyle\xi\triangleq\max_{n,w_{\mathcal{V}_n},i,i'}\left|\left[\left(\overline{\Sigma}_{Y_{w_{\mathcal{V}_n}} Y_{w_{\mathcal{V}_n}}}\right)^{\hspace{-1mm}-1} \right]_{ii'}\right|
\label{max1}\vspace{-1mm}
\end{equation}
and $
\epsilon\triangleq\displaystyle 0.5\log 1\hspace{-0.5mm}\Big{/}\hspace{-1.5mm}\left(1\hspace{-0.5mm}-\hspace{-0.5mm}\left(K^{1.5}L^{2.5}\kappa\xi\varepsilon\right)^2\right)
$.
In the result below, we prove that the joint walk $\widehat{w}\triangleq(\widehat{w}_{\mathcal{V}_1},\ldots, \widehat{w}_{\mathcal{V}_{\mathcal{K}}})$ 
is guaranteed to achieve an entropy $\overline{\mathbb{H}}\hspace{-0.5mm}\left[Z_{Y_{\widehat{w}}}\right]$ (i.e., by plugging $\widehat{w}$ into (\ref{maxent21})) that is not more than $\epsilon$ 
from the maximum entropy $\overline{\mathbb{H}}\hspace{-0.5mm}\left[Z_{Y_{w^{\ast}}}\right]$ achieved by joint walk $w^{\ast}$ (\ref{maxent2}):\vspace{-1mm}
\begin{thm}[Performance Guarantee]
If $K^{1.5}L^{2.5}\kappa\xi\varepsilon\hspace{-0.06mm}<1$, then
$\overline{\mathbb{H}}\hspace{-0.5mm}\left[Z_{Y_{w^{\ast}}}\right] - \overline{\mathbb{H}}\hspace{-0.5mm}\left[Z_{Y_{\widehat{w}}}\right]\leq\epsilon$.
\label{thm2}\vspace{-1mm}
\end{thm}
\ifthenelse{\value{sol}=1}{The proof of Theorem~\ref{thm2} is given in Appendix~\ref{equ2}.} 
{Its proof is given in \citep{LowArxiv12}.} 
The implication of Theorem~\ref{thm2} is that our partially decentralized active sensing algorithm can perform comparatively well (i.e., small $\epsilon$) under the following favorable environmental conditions: (a) the network of $K$ sensors is not large, (b) length $L$ of each sensor's walk to be optimized is not long, (c) the largest subset of $\kappa$ sensors being formed to coordinate their walks (i.e., largest connected component in $\mathcal{G}$) is reasonably small, and (d) the minimum required correlation $\varepsilon$ between walks of adjacent sensors is kept low.\vspace{-0mm}

Algorithm~\ref{dadgf} below outlines the key operations of our D$^2$FAS algorithm to be run on each mobile sensor $k$, as detailed previously in Sections~\ref{sec:ddf} and~\ref{sec:dadgf}:\vspace{-3mm}
%
\begin{algorithm}
{\scriptsize
\DontPrintSemicolon
\While{true}
{
\tcc{Data fusion (Section~\ref{sec:ddf})}
Construct local summary by (\ref{lsf}) \& (\ref{lsc})\;    
Exchange local summary with every sensor $i\neq k$\; 
Construct global summary by (\ref{gsf}) \& (\ref{gsc})\; 
Predict measurements at unobserved road segments by (\ref{mupitc}) \& (\ref{varpitc})\;
\tcc{Active Sensing (Section~\ref{sec:dadgf})}    
Construct $\Phi_k$ by (\ref{phik})\; 
Exchange $\Phi_k$ with every sensor $i\neq k$\;    
Compute adjacency vector $a_k$ by (\ref{rel2}) \& (\ref{adjv})\;
Exchange adjacency vector with every sensor $i\neq k$\;  
Construct adjacency matrix of coordination graph\;
Find vertex set $\mathcal{V}_n$ of its residing connected component\;
Compute maximum-entropy joint walk $\widehat{w}_{\mathcal{V}_n}$ by (\ref{dmaxent})\;
Execute walk $\widehat{w}_k$ and observe its road segments $Y_{\widehat{w}_k}$\;
Update local information $D_k$, $z_{D_k}$, and $s_k$ by (\ref{upDk})\; 
}
\caption{D$^2$FAS$(U, K, L, k, D_k, z_{D_k}, s_k)$}
  \label{dadgf}
}
\end{algorithm}
\vspace{-7mm}
\section{Time and Communication Overheads}
\label{overhead}\vspace{-3mm}
In this section, the time and communication overheads of our D$^2$FAS algorithm are analyzed and
compared to that of centralized active sensing (\ref{maxent2}) coupled with the data fusion methods: Full GP (FGP) and SoD (Section~\ref{sec:gpgk}).\vspace{-2mm}
%
%
\subsection{Time Complexity}
\label{sec:time}\vspace{-2mm}
%
The data fusion component of D$^2$FAS involves computing the local and global summaries and the predictive Gaussian distribution.
To construct the local summary using (\ref{lsf}) and (\ref{lsc}), each sensor has to evaluate $\Sigma_{D_k D_k|U}$ in $\set{O}\hspace{-1mm}\left(|U|^3 + |U| (|D|/K)^2\right)$ time and invert it in $\set{O}\hspace{-1mm}\left((|D|/K)^3\right)$ time, after which the local summary is obtained in $\set{O}\hspace{-1mm}\left(|U|^2|D|/K + |U|(|D|/K)^2\right)$ time. 
The global summary is computed in $\set{O}\hspace{-1mm}\left(|U|^2 K\right)$ by (\ref{gsf}) and (\ref{gsc}). 
Finally, the predictive Gaussian distribution is derived in $\set{O}\hspace{-1mm}\left(|U|^3 + |U||Y|^2 \right)$ time using (\ref{mupitc}) and (\ref{varpitc}).
Supposing $|Y|\leq |U|$ for simplicity, the time complexity of data fusion is then $\set{O}\hspace{-1mm}\left((|D|/K)^3 + |U|^3 + |U|^2 K\right)$.  

Let the maximum out-degree of $G$ be denoted by $\delta$. Then, each sensor has to consider $\Delta\triangleq\delta^L$ possible walks of length $L$.
The active sensing component of D$^2$FAS involves computing $\Phi_k$ in $\set{O}\hspace{-1mm}\left(\Delta L |U|^2\right)$ time, $a_k$ in $\set{O}\hspace{-1mm}\left(\Delta^{2} L^2 |U| K\right)$ time, its residing connected component in $\set{O}\hspace{-1mm}\left(\kappa^2\right)$ time, and the maximum-entropy joint walk by (\ref{varpitc}) and (\ref{dmaxent}) with the following incurred time: 
The largest connected component of $\kappa$ sensors in $\mathcal{G}$ has to consider $\Delta^{\kappa}$ possible joint walks. Note that 
$\overline{\Sigma}_{Y_{w_{\mathcal{V}_n}} Y_{w_{\mathcal{V}_n}}} = \mbox{diag}\hspace{-1mm}\left((\Sigma_{Y_{w_k}Y_{w_k}|U})_{k\in \mathcal{V}_n}\right) + \Sigma_{Y_{w_{\mathcal{V}_n}}U}\ddot{\Sigma}_{UU}^{-1}\Sigma_{UY_{w_{\mathcal{V}_n}}}$ where $\mbox{diag}(B)$ constructs a diagonal matrix by placing vector $B$ on its diagonal. By exploiting $\Phi_k$, the diagonal and latter matrix terms for all possible joint walks can be computed in $\set{O}\hspace{-1mm}\left(\kappa\Delta(L |U|^2 + L^2 |U|)\right)$ and
$\set{O}\hspace{-1mm}\left(\kappa^2\Delta^{2}L^2 |U|\right)$ time, respectively.
For each joint walk $w_{\mathcal{V}_n}$, evaluating the determinant of $\overline{\Sigma}_{Y_{w_{\mathcal{V}_n}} Y_{w_{\mathcal{V}_n}}}$ incurs $\set{O}\hspace{-1mm}\left((\kappa L)^3 \right)$ time.
Therefore, the time complexity of active sensing is $\set{O}\hspace{-1mm}\left(\kappa\Delta L |U|^2 + \Delta^{2} L^2 |U| (K+\kappa^2) +\Delta^{\kappa} (\kappa L)^3\right)$. 

Hence, the time complexity of our D$^2$FAS algorithm is $\set{O}\hspace{-0mm}((|D|/K)^3 + |U|^2 (|U| + K + \kappa\Delta L) + \Delta^{2} L^2 |U| (K+\kappa^2)+ \Delta^{\kappa}(\kappa L)^3)$. In contrast, the time incurred by centralized active sensing coupled with FGP and SoD are, respectively, $\set{O}\hspace{-1mm}\left(|D|^3+\Delta^{K}KL(|D|^2 + (KL)^2)\right)$ and $\set{O}\hspace{-1mm}\left(|U|^3|D|+\Delta^{K}KL(|U|^2 + (KL)^2)\right)$. It can be observed that D$^2$FAS can scale better with large $|D|$ (i.e., number of observations) and $K$ (i.e., number of sensors). The scalability of D$^2$FAS vs. FGP and SoD will be further evaluated empirically in Section~\ref{sec:exp}.\vspace{-2mm}
%
%
\subsection{Communication Complexity}
\label{sec:comm}\vspace{-2mm}
Let the communication overhead be defined as the size of each broadcast message.
Recall from the data fusion component of D$^2$FAS in Algorithm~\ref{dadgf} that, in each iteration, each sensor broadcasts a $\set{O}\hspace{-1mm}\left(|U|^2\right)$-sized summary encapsulating its local observations, which is robust against communication failure. 
In contrast,
FGP and SoD require each sensor to broadcast, in each iteration, a $\set{O}\hspace{-1mm}\left(|D|/K\right)$-sized message comprising exactly its local observations to handle communication failure.
If the number of local observations grows to be larger in size than a local summary of predefined size, then the data fusion component of D$^2$FAS is more scalable than FGP and SoD in terms of communication overhead.
For the partially decentralized active sensing component of D$^2$FAS, each sensor broadcasts $\set{O}\hspace{-1mm}\left(\Delta L|U| \right)$-sized $\Phi_k$ and $\set{O}\hspace{-1mm}\left(K\right)$-sized $a_k$ messages.
%
%
\section{Experiments and Discussion}
\label{sec:exp}\vspace{-3mm}
\begin{wrapfigure}{l}{4.1cm}
\vspace{-4mm}\includegraphics[scale=0.40]{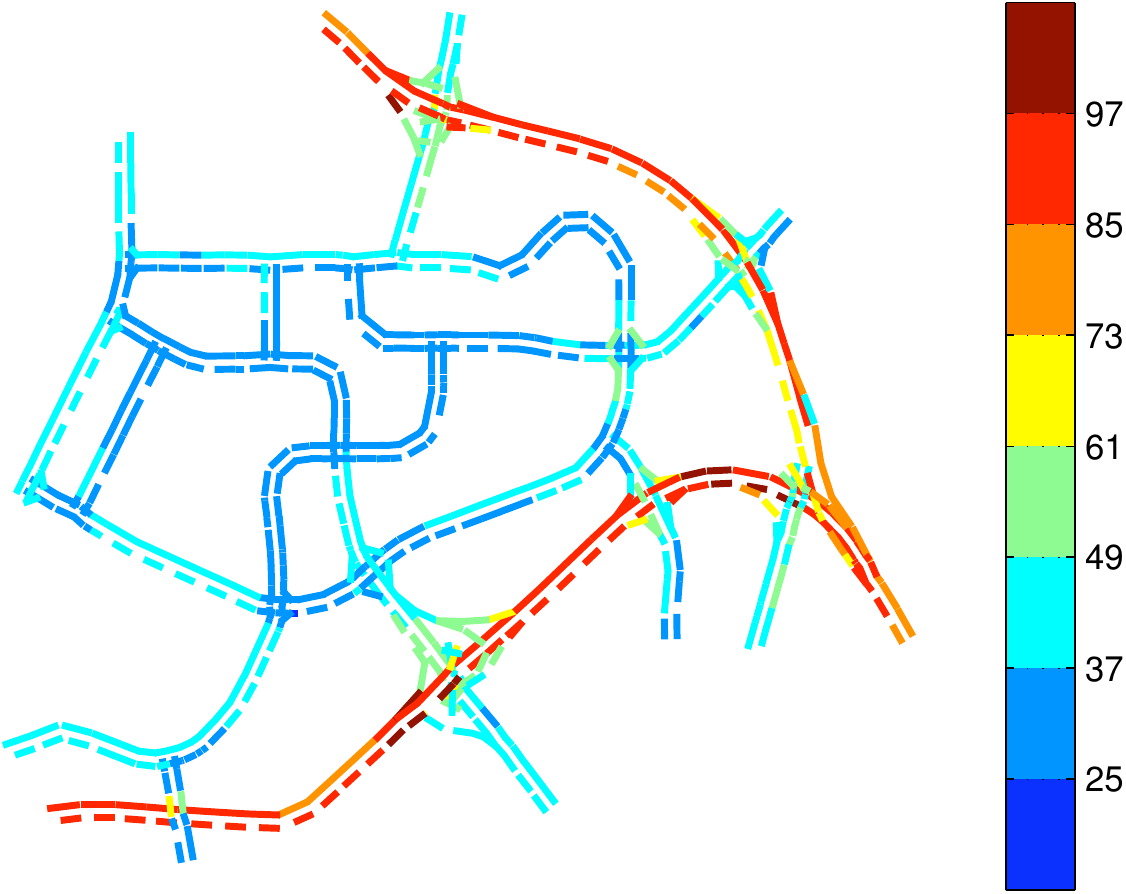}\vspace{-4mm}
\end{wrapfigure}
This section evaluates the predictive performance, time efficiency, and scalability of our D$^2$FAS algorithm on a real-world traffic phenomenon (i.e., speeds (km/h) of road segments) over an urban road network
(top figure) in Tampines area, Singapore during evening peak hours on April $20$, $2011$.
It comprises $775$ road segments including highways, arterials, slip roads, etc. The mean speed is $48.8$ km/h and the standard deviation is $20.5$ km/h.

The performance of D$^2$FAS is compared to that of centralized active sensing (\ref{maxent2}) coupled with the state-of-the-art data fusion methods: full GP (FGP) and SoD (Section~\ref{sec:gpgk}).  
A network of $K$ mobile sensors is tasked to explore the road network to gather a total of up to $960$ observations. 
To reduce computational time, each sensor repeatedly computes and executes maximum-entropy walks of length $L=2$ (instead of computing a very long walk), unless otherwise stated. For D$^2$FAS and SoD, $|U|$ is set to $64$ .  
For the active sensing component of D$^2$FAS, $\varepsilon$ is set to $0.1$, unless otherwise stated.
The experiments are run on a Linux PC with Intel$\circledR$ Core$^{\mbox{\tiny{TM}}}2$ Quad CPU Q$9550$ at $2.83$~GHz.\vspace{-3mm}
%
%
\subsection{Performance Metrics}\vspace{-3mm}
%
The first metric evaluates the predictive performance of a tested algorithm: It measures the \emph{root mean squared error} (RMSE)
%
$\sqrt{|V|^{-1}\sum_{s\in V}\left(z_s-\widehat{\mu}_s\right)^2}$
%
over the entire domain $V$ of the road network that is incurred by the predictive mean $\widehat{\mu}_s$ of the tested algorithm, specifically, using (\ref{pmu}) of FGP,
(\ref{eq:fsod}) of SoD, or (\ref{mupitc}) of D$^2$FAS.
The second metric evaluates the time efficiency and scalability of a tested algorithm by measuring its incurred time; for D$^2$FAS, the maximum of the time incurred by all subsets $\mathcal{V}_1,\ldots,\mathcal{V}_{\mathcal{K}}$ of sensors is recorded.\vspace{-3mm}
%
%
\subsection{Results and Analysis}
\label{ppte}\vspace{-3mm}
\begin{figure}[h]
\begin{tabular}{ccc}
\hspace{-2mm}\includegraphics[scale=0.143]{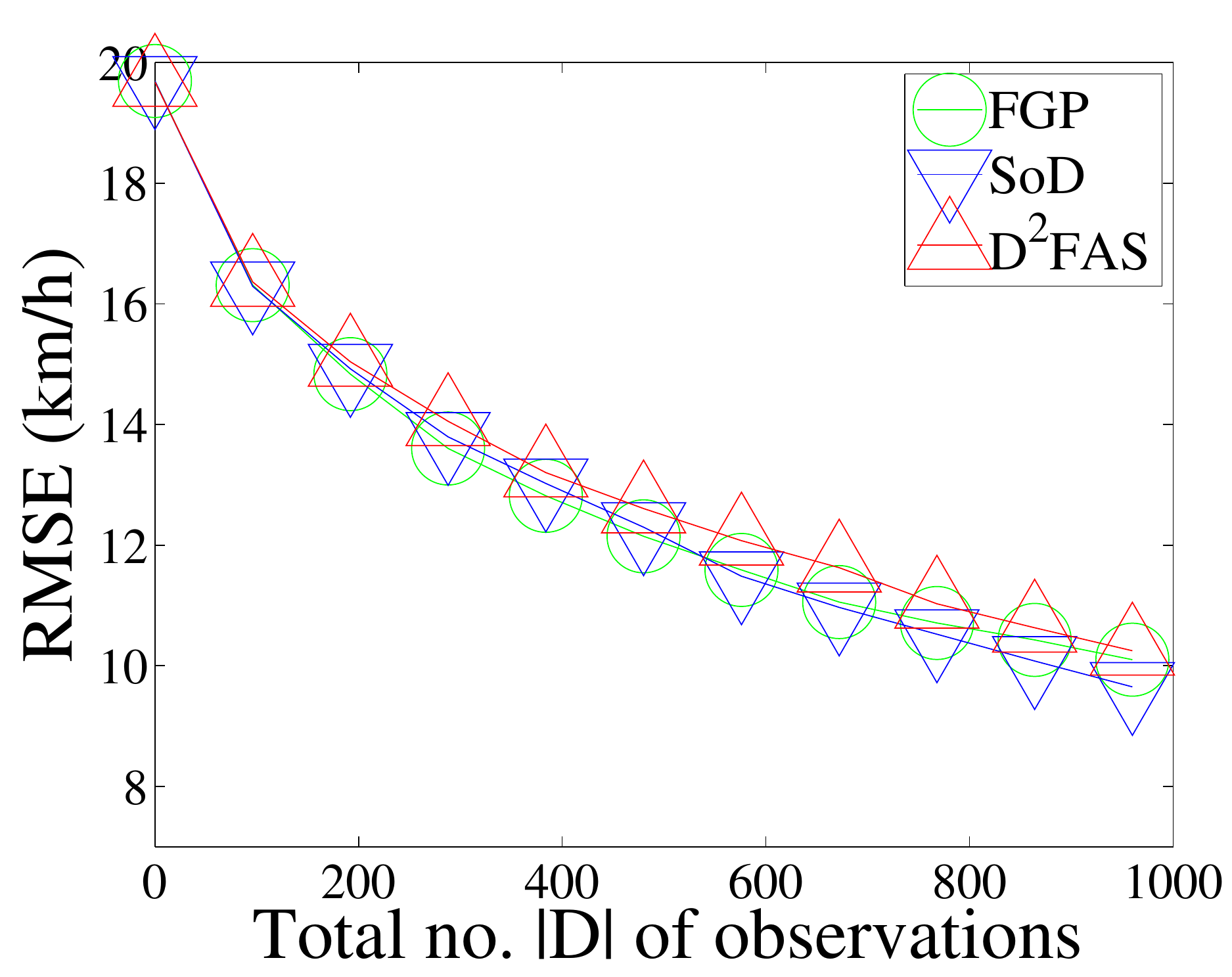} & \hspace{-4.5mm}\includegraphics[scale=0.143]{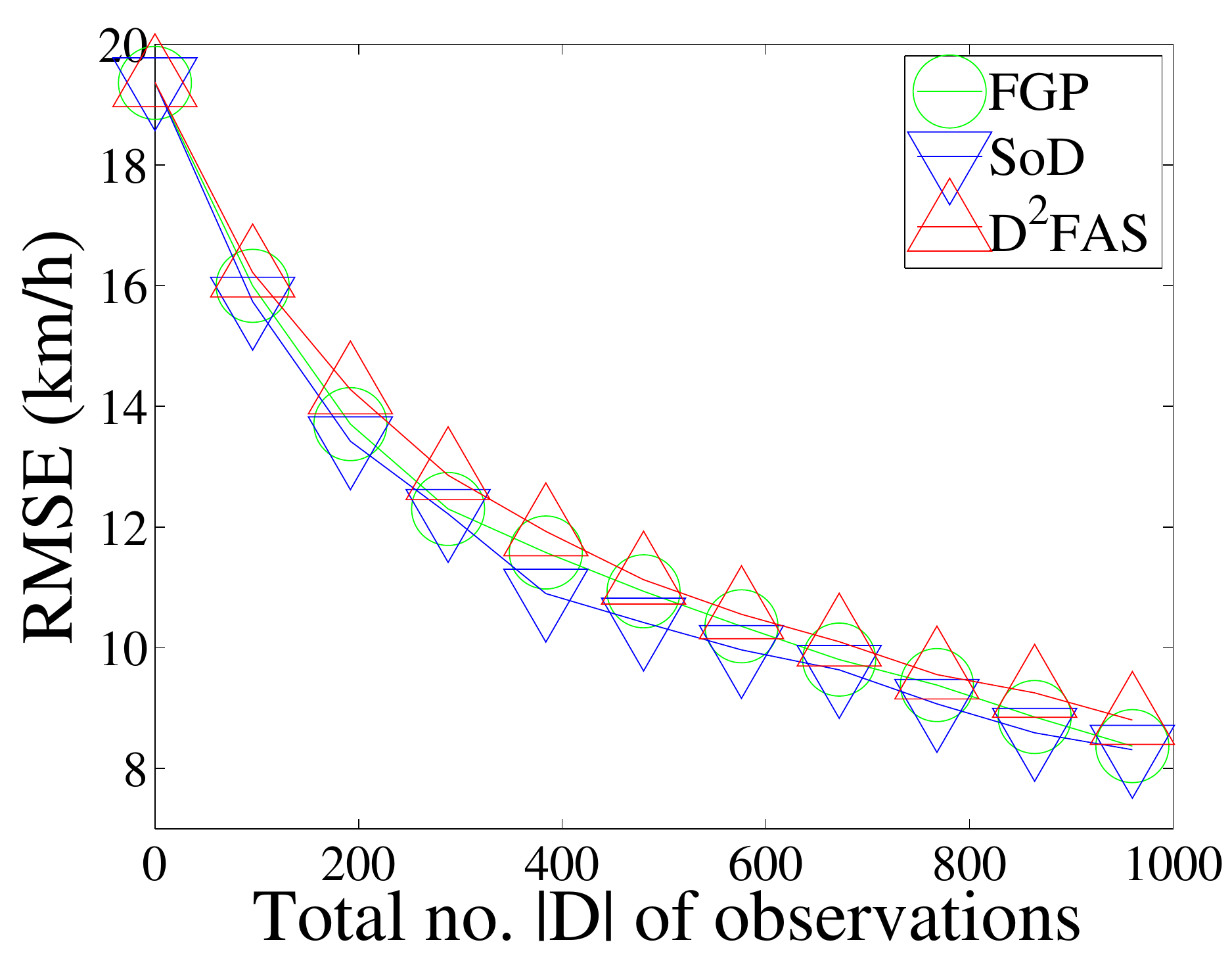} &  \hspace{-4.5mm}\includegraphics[scale=0.143]{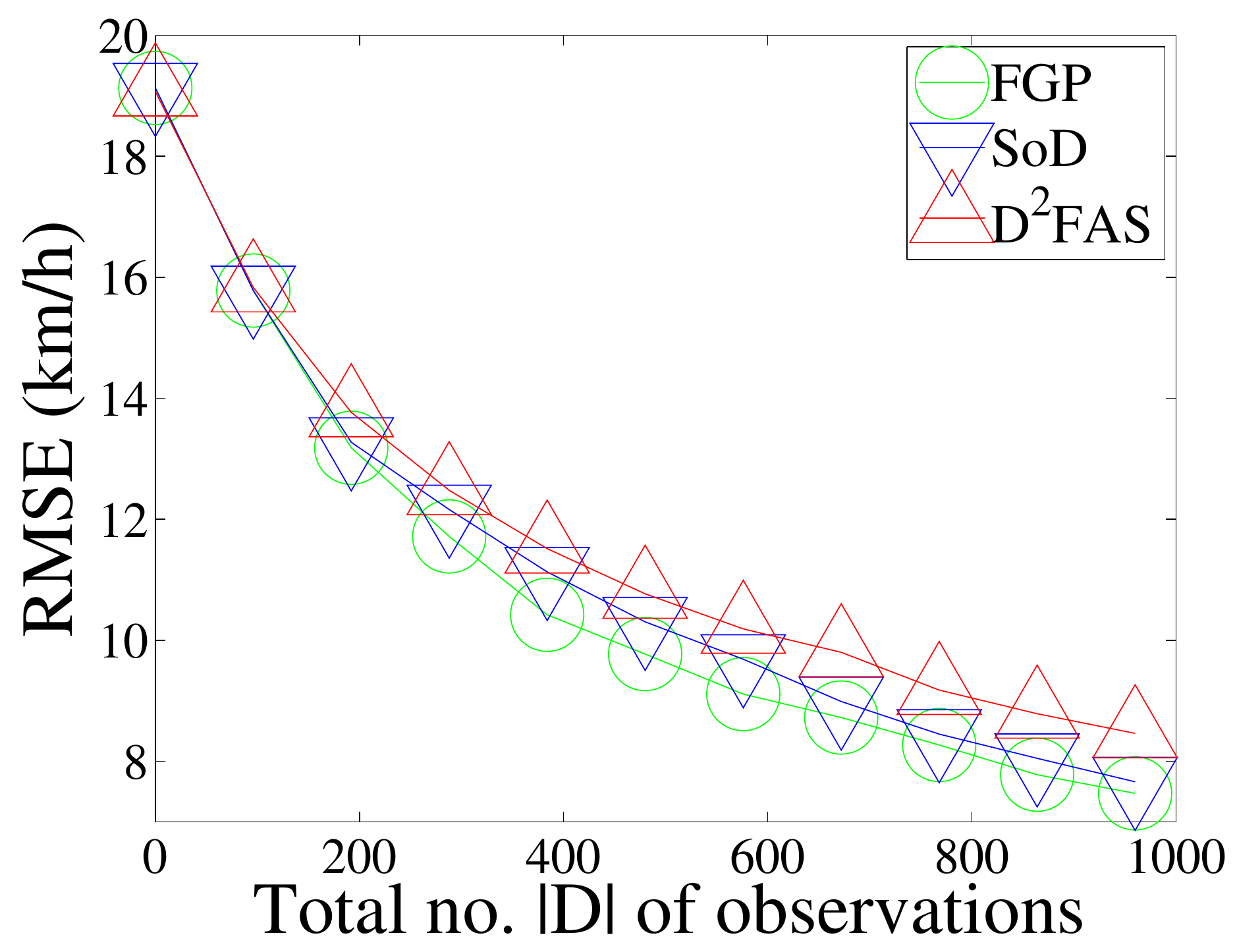}\vspace{-1mm}\\
\hspace{-2mm}(a) $K = 4$ & \hspace{-4.5mm}(b) $K = 6$ &  \hspace{-4.5mm}(c) $K = 8$\\ 
 \hspace{-2mm}\includegraphics[scale=0.143]{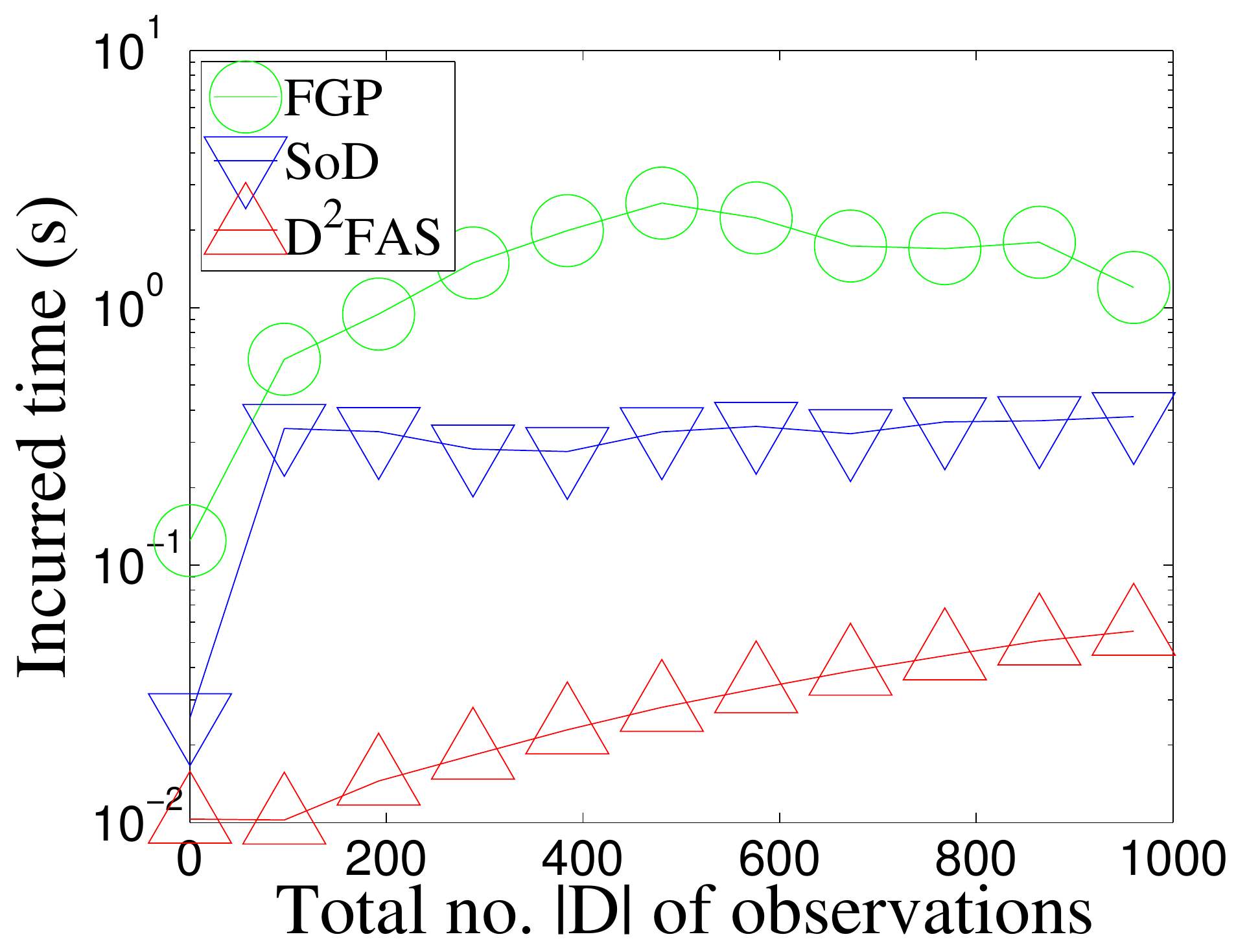} & \hspace{-4.5mm}\includegraphics[scale=0.143]{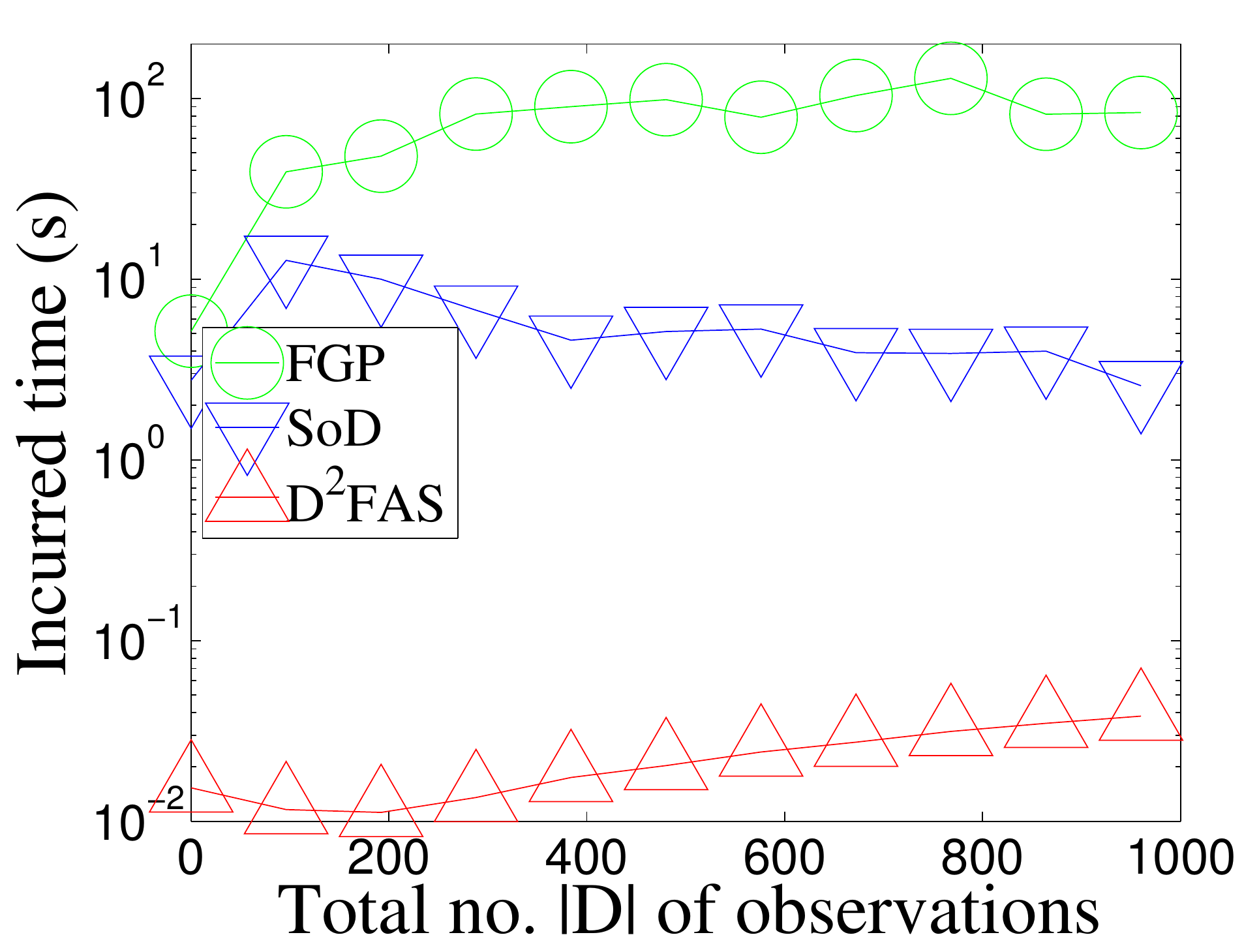} &  \hspace{-4.5mm}\includegraphics[scale=0.143]{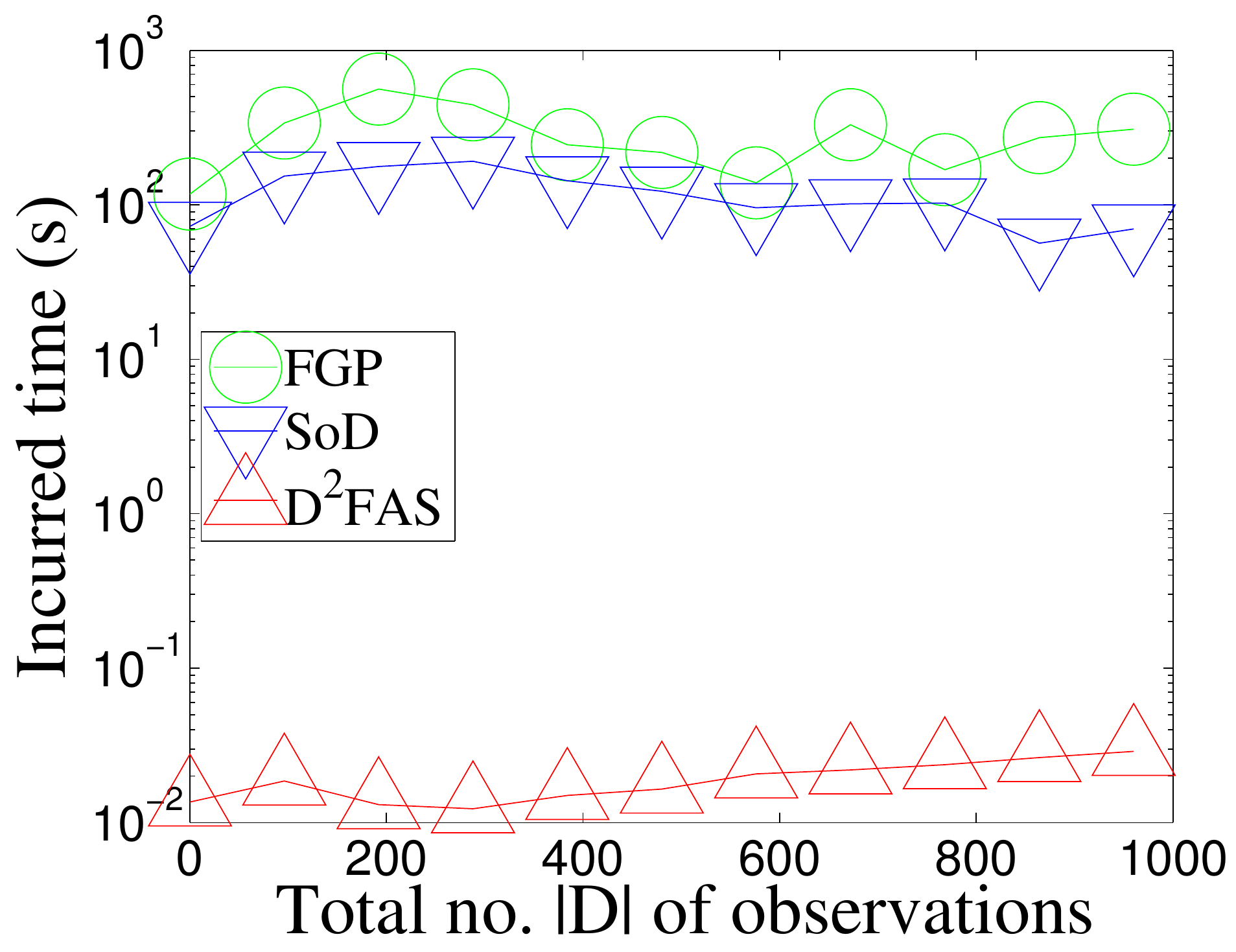}\vspace{-1mm}\\
\hspace{-2mm}(d) $K = 4$ & \hspace{-4.5mm}(e) $K = 6$ &  \hspace{-4.5mm}(f) $K = 8$\vspace{-3mm} 
\end{tabular}
\caption{Graphs of (a-c) predictive performance and (d-f) time efficiency vs. total no. $|D|$ of observations gathered by varying number $K$ of mobile sensors.}
\label{fig3}\vspace{-2mm}
\end{figure}
{\bf Predictive performance and time efficiency.} 
Fig.~\ref{fig3} shows results of the performance of the tested algorithms averaged over $40$ randomly generated starting sensor locations with varying number $K=4, 6, 8$ of sensors. It can be observed that D$^2$FAS is significantly more time-efficient 
and scales better with increasing number $|D|$ of observations 
(Figs.~\ref{fig3}d to~\ref{fig3}f) while achieving predictive performance close to that of centralized active sensing coupled with FGP and SoD (Figs.~\ref{fig3}a to~\ref{fig3}c).
Specifically, D$^2$FAS is about $1, 2, 4$ orders of magnitude faster than centralized active sensing coupled with FGP and SoD for $K = 4,6,8$ sensors, respectively.
%
%
\begin{figure}
\begin{tabular}{ccc}
\hspace{-2.5mm}\includegraphics[scale=0.143]{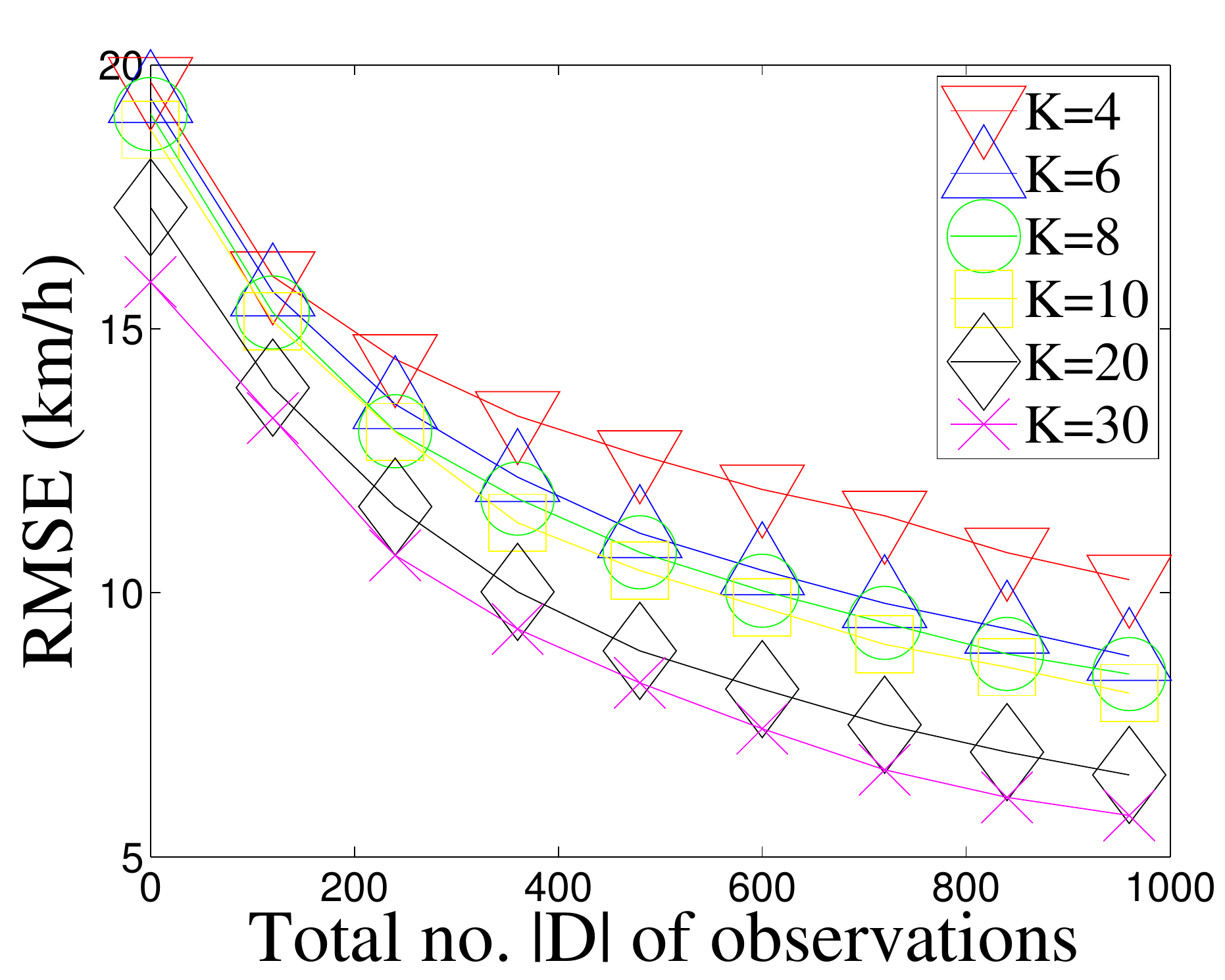} & \hspace{-4mm}\includegraphics[scale=0.143]{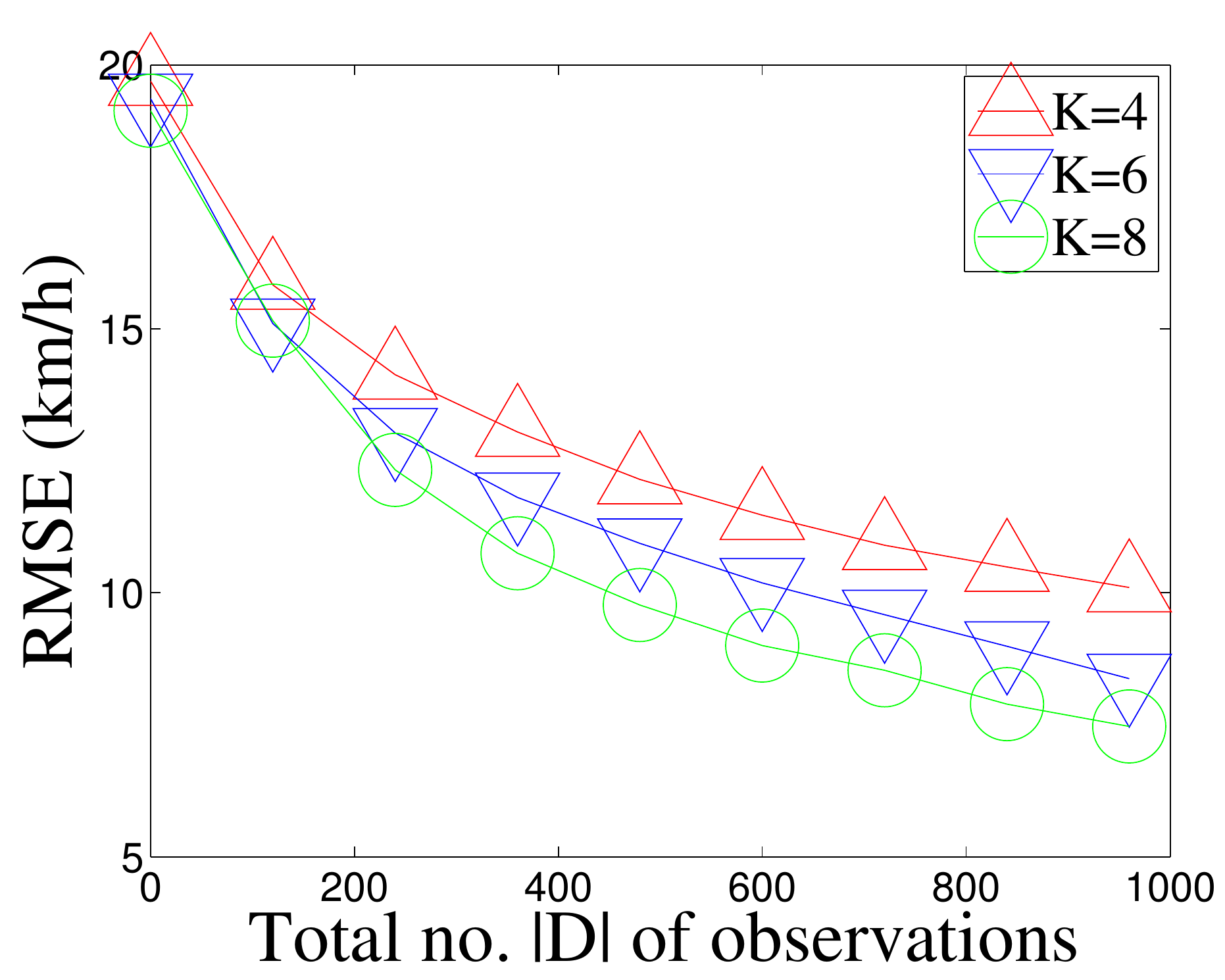} &  \hspace{-4mm}\includegraphics[scale=0.143]{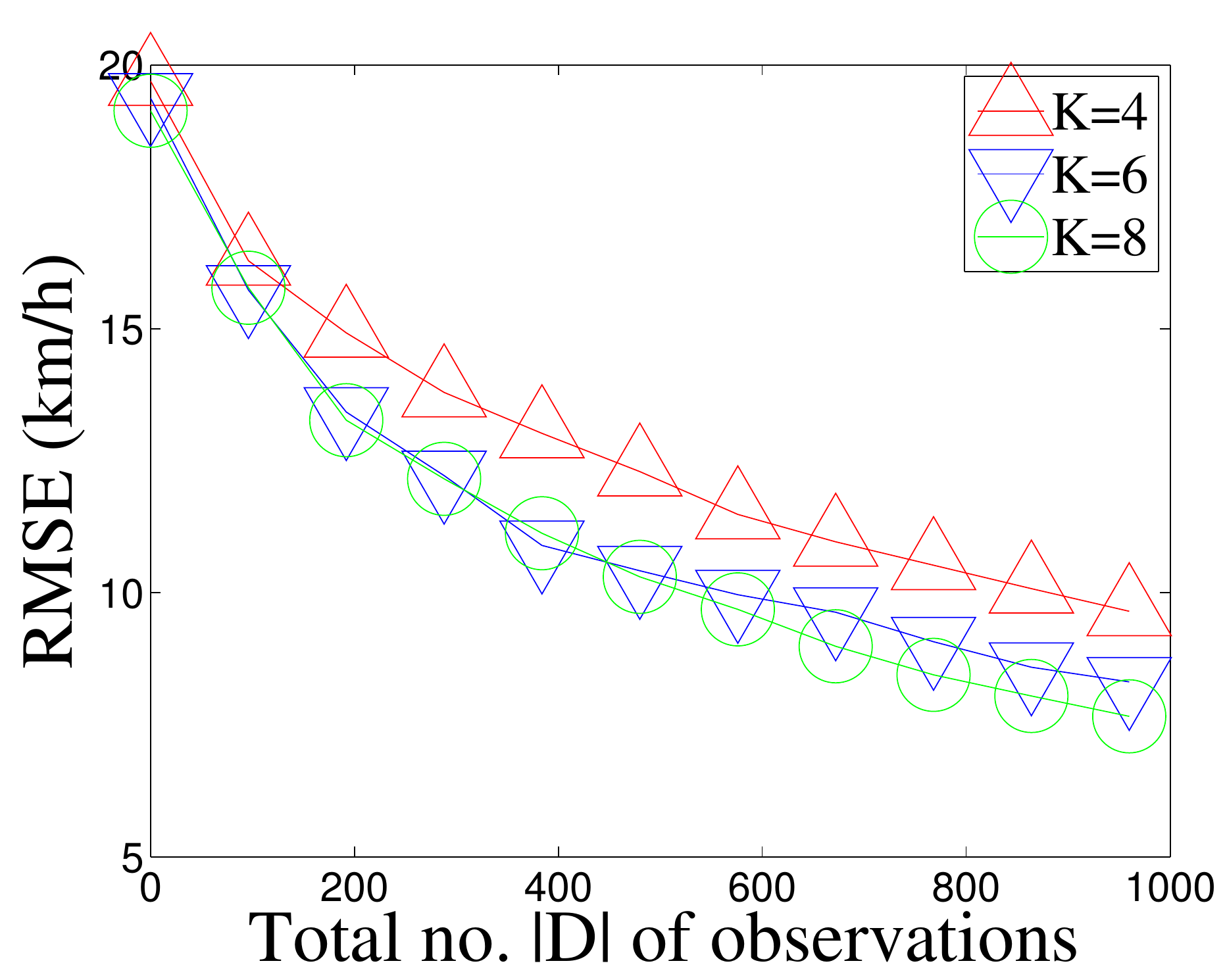}\vspace{-1mm}\\
\hspace{-2.5mm}(a) D$^2$FAS & \hspace{-4mm}(b) FGP &  \hspace{-4mm}(c) SoD\\ 
 \hspace{-2.5mm}\includegraphics[scale=0.143]{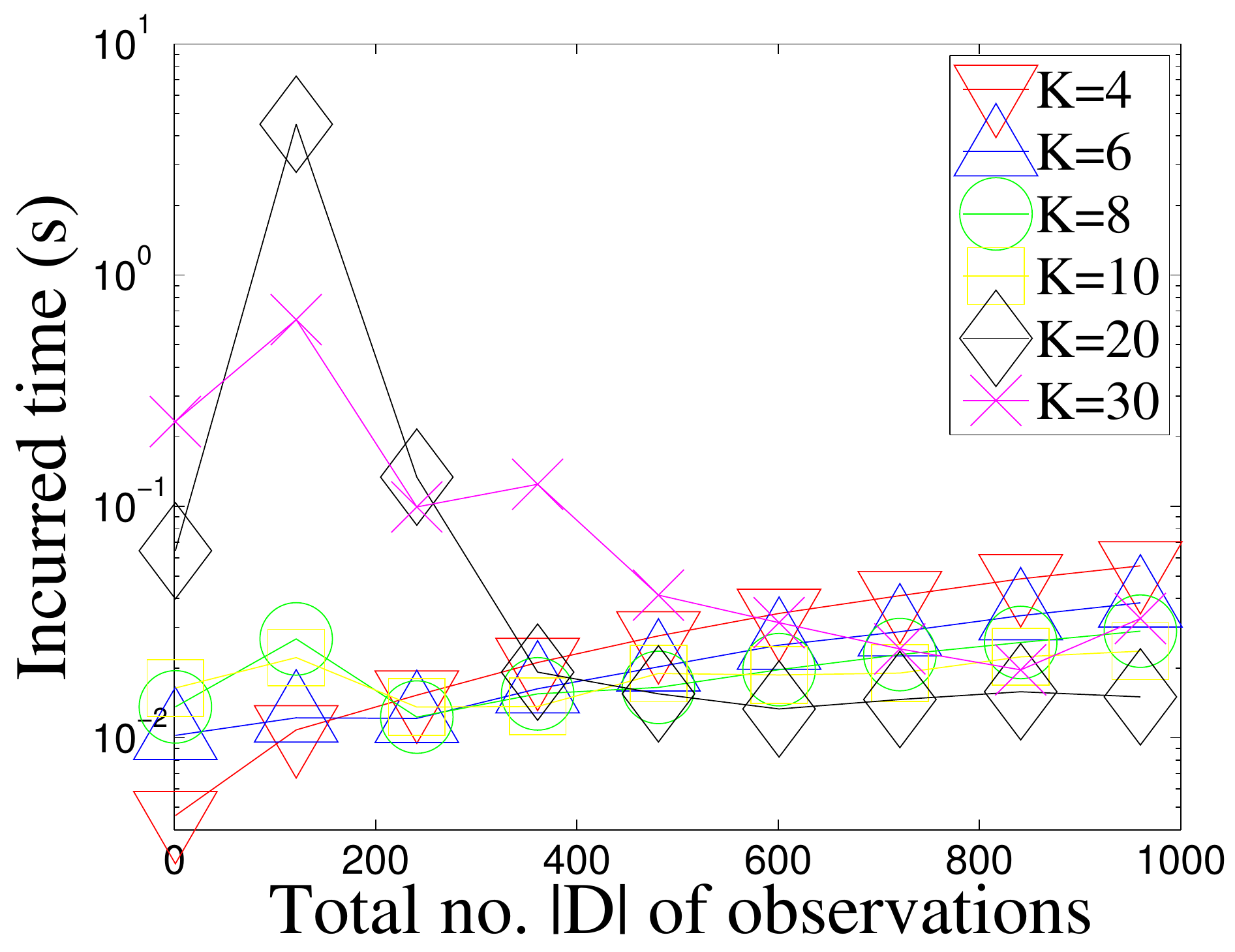} & \hspace{-4mm}\includegraphics[scale=0.143]{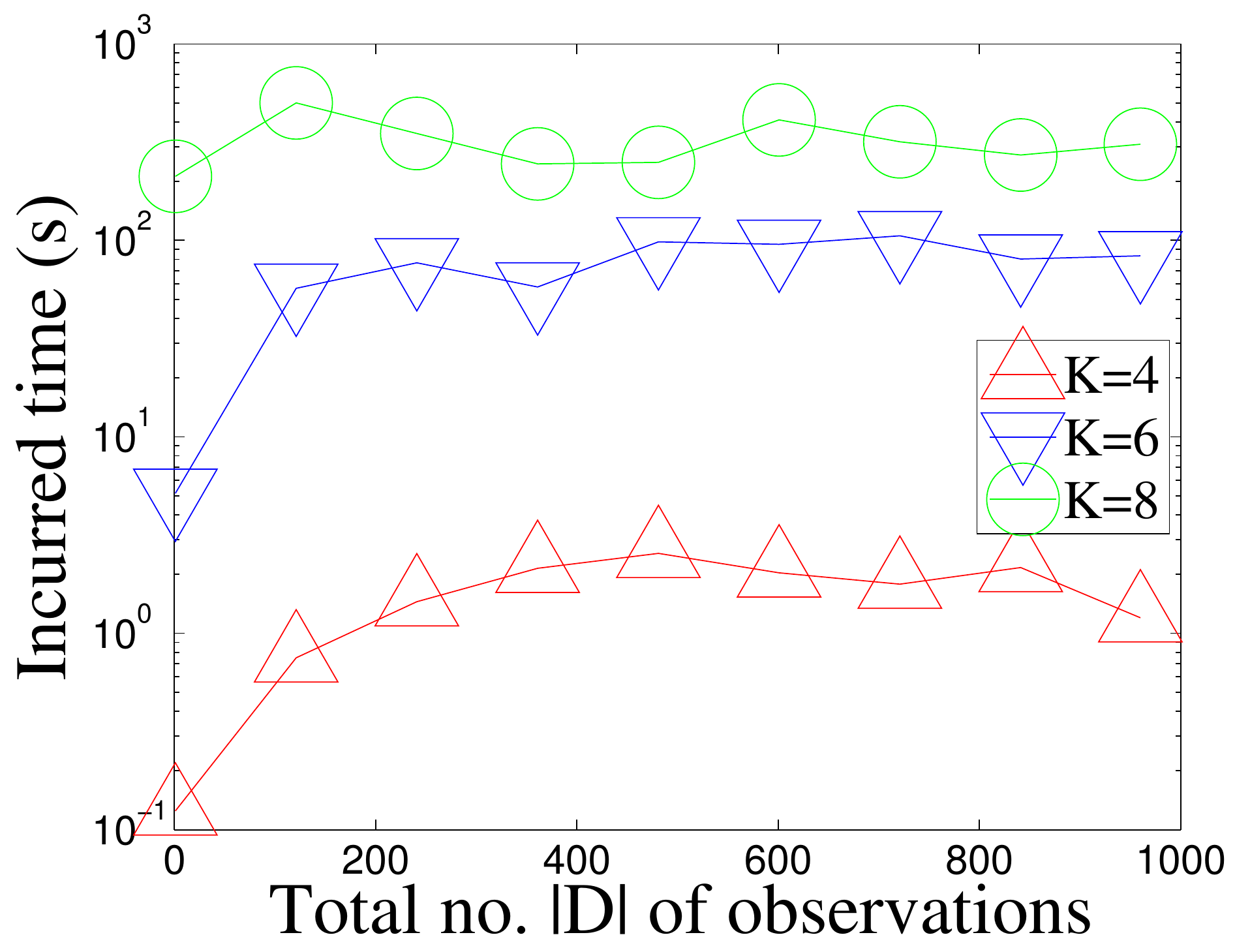} &  \hspace{-4mm}\includegraphics[scale=0.143]{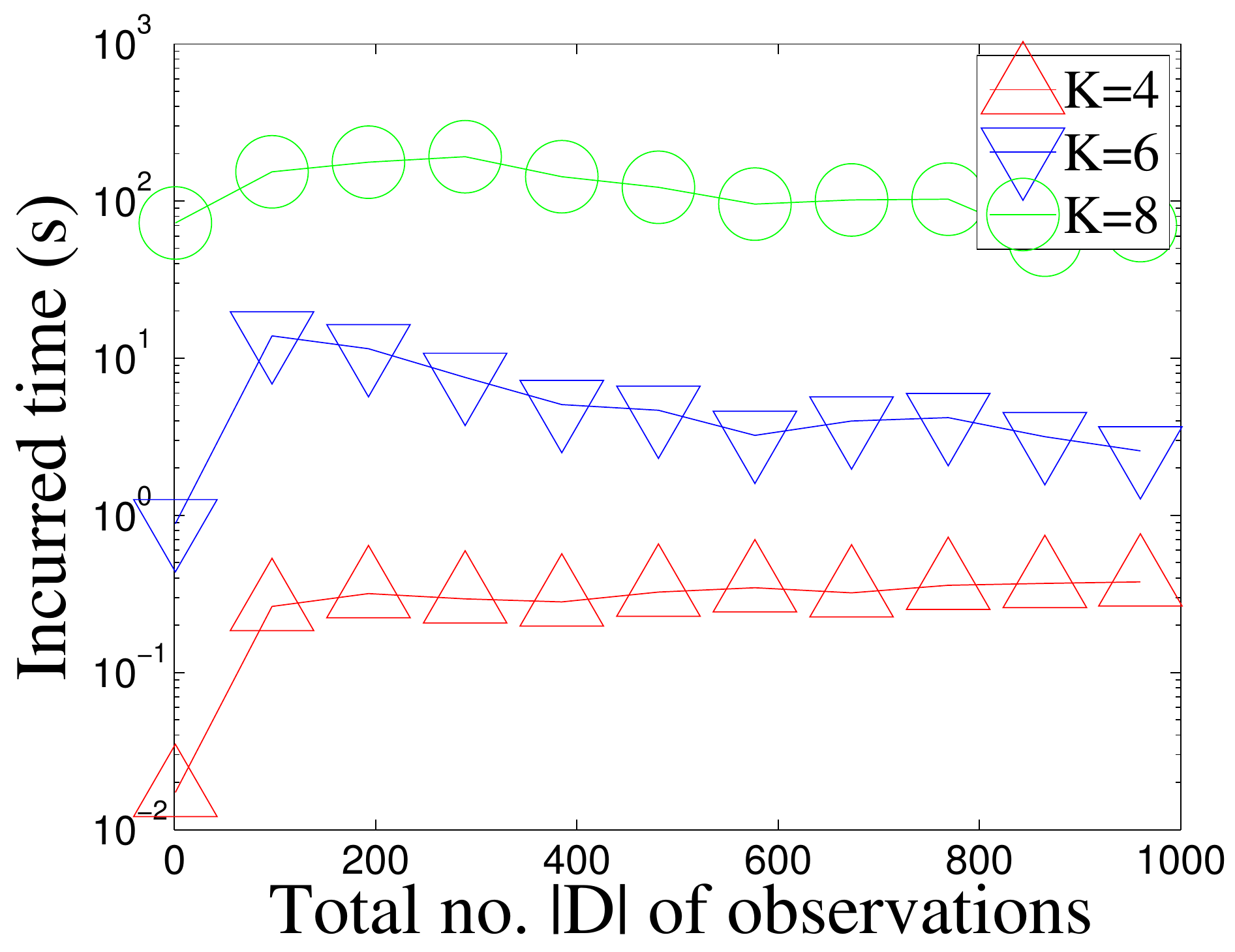}\vspace{-1mm}\\
\hspace{-2.5mm}(d) D$^2$FAS & \hspace{-4mm}(e) FGP &  \hspace{-4mm}(f) SoD\vspace{-3mm} 
\end{tabular}
\caption{Graphs of (a-c) predictive performance and (d-f) time efficiency vs. total no. $|D|$ of observations gathered by varying number $K$ of mobile sensors.}
\label{fig6}\vspace{-4mm}
\end{figure}

{\bf Scalability of D$^2$FAS.} Using the same results as that in Fig.~\ref{fig3}, Fig.~\ref{fig6} plots them differently to reveal the scalability of the tested algorithms with increasing number $K$ of sensors.
Additionally, we provide results of the performance of D$^2$FAS for $K=10, 20, 30$ sensors; 
such results are not available for centralized active sensing coupled with FGP and SoD due to extremely long incurred time.
It can be observed from Figs.~\ref{fig6}a to~\ref{fig6}c that the predictive performance of all tested algorithms improve with a larger number of sensors because each sensor needs to execute fewer walks and its performance is therefore less adversely affected by its myopic selection (i.e., $L=2$) of maximum-entropy walks. As a result, more informative unobserved road segments are explored.

As shown in Fig.~\ref{fig6}d, when the randomly placed sensors gather their initial observations (i.e., $|D|<400$), the time incurred by D$^2$FAS is higher for greater $K$ due to larger subsets of sensors being formed to coordinate their walks (i.e., larger $\kappa$). As more observations are gathered (i.e., $|D|\geq 400$), its partially decentralized active sensing component directs the sensors to explore further apart from each other in order to maximize the entropy of their walks.
This consequently decreases $\kappa$, leading to a reduction in incurred time.
Furthermore, as $K$ increases from $4$ to $20$, the incurred time decreases due to its decentralized data fusion component that can distribute the computational load among a greater number of sensors.
When the road network becomes more crowded from $K=20$ to $K=30$ sensors, the incurred time increases slightly due to slightly larger $\kappa$.
In contrast, Figs.~\ref{fig6}e and~\ref{fig6}f show that the time taken by FGP and SoD increases significantly primarily due to their centralized active sensing incurring exponential time in $K$.
Hence, the scalability of our D$^2$FAS algorithm in the number of sensors allows the deployment of a larger-scale mobile sensor network (i.e., $K\geq 10$) to achieve more accurate traffic modeling and prediction (Figs.~\ref{fig6}a to~\ref{fig6}c).
\begin{figure}
\begin{tabular}{ccc}
 \hspace{-2mm}\includegraphics[scale=0.143]{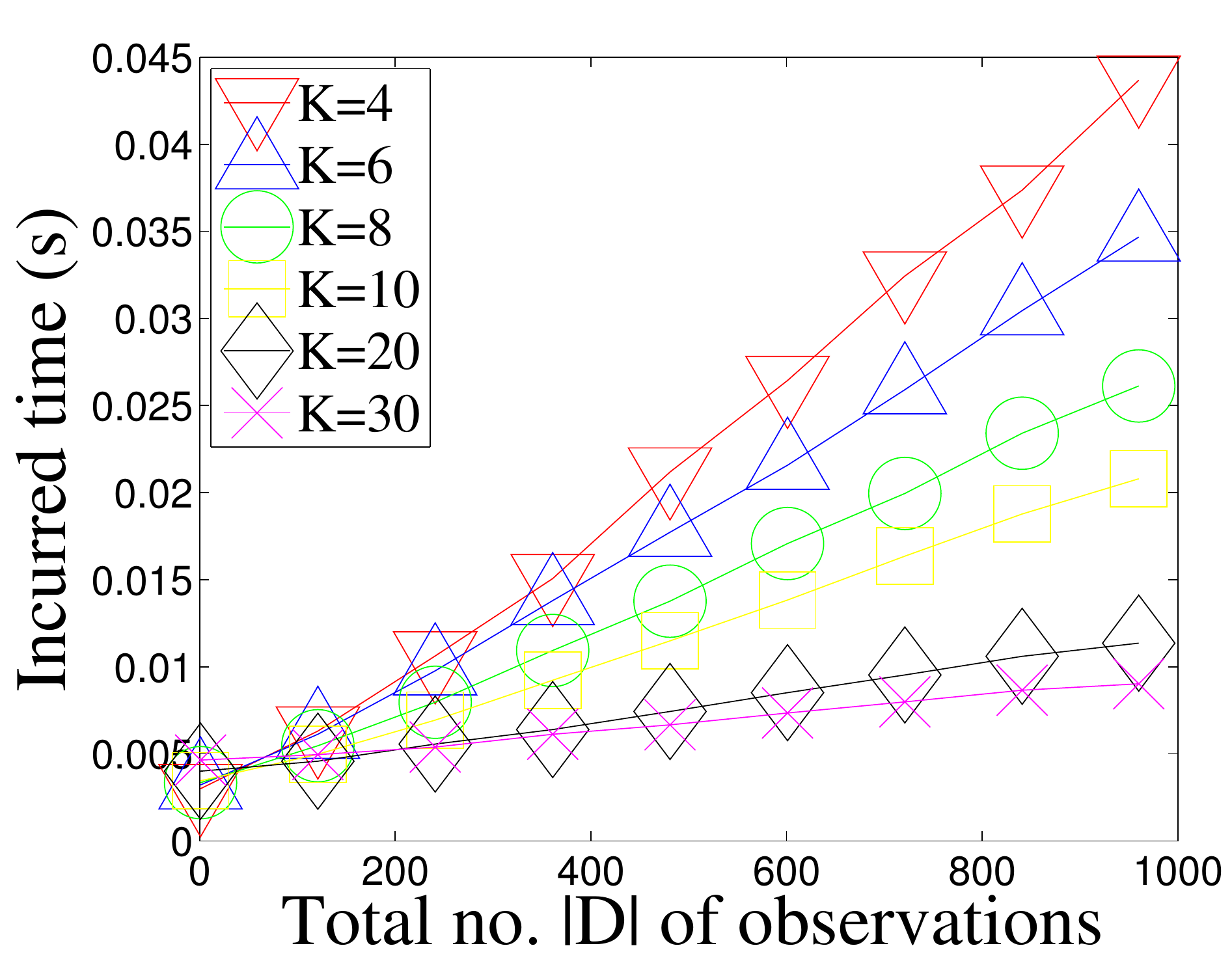} & \hspace{-4mm}\includegraphics[scale=0.143]{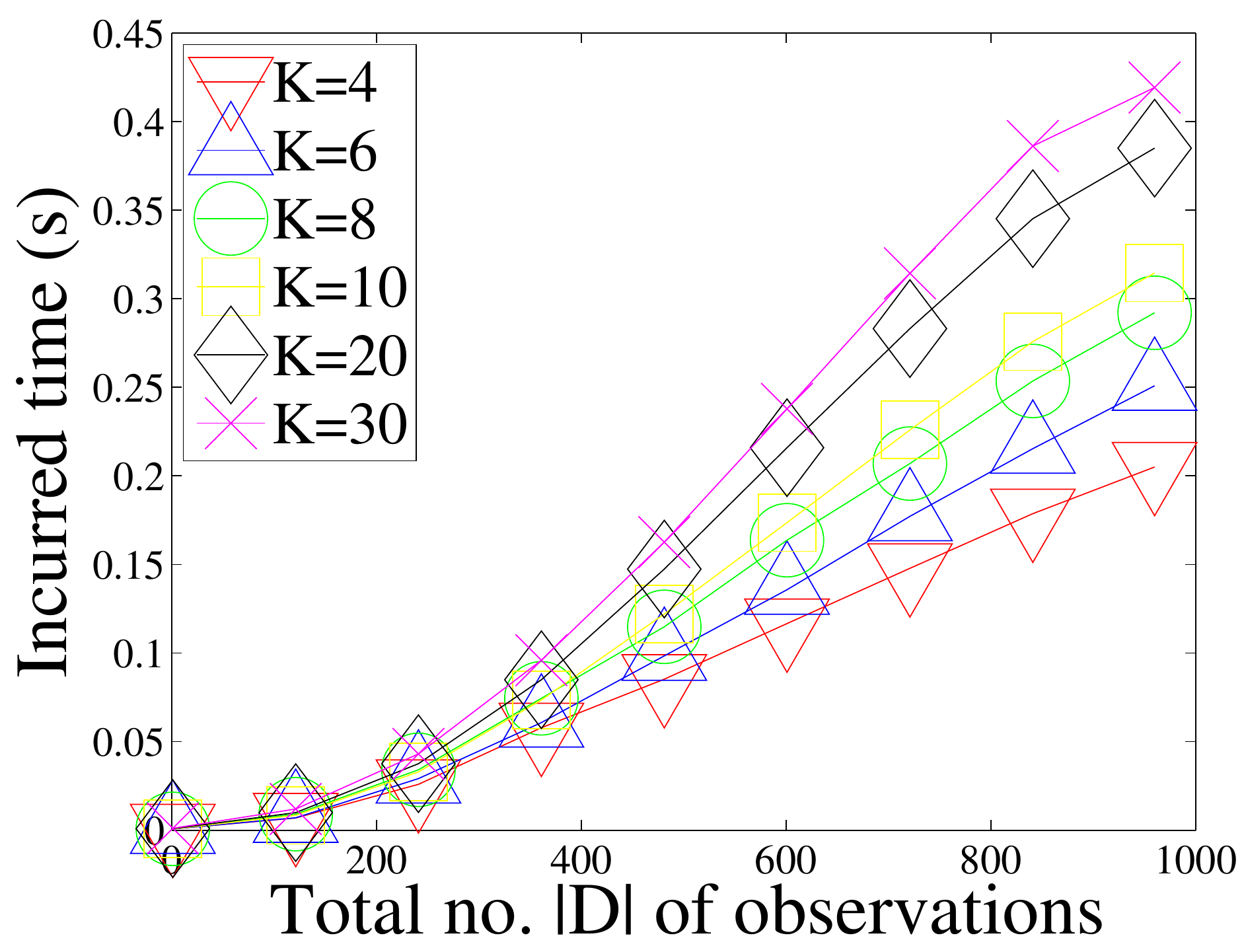} &  \hspace{-4mm}\includegraphics[scale=0.143]{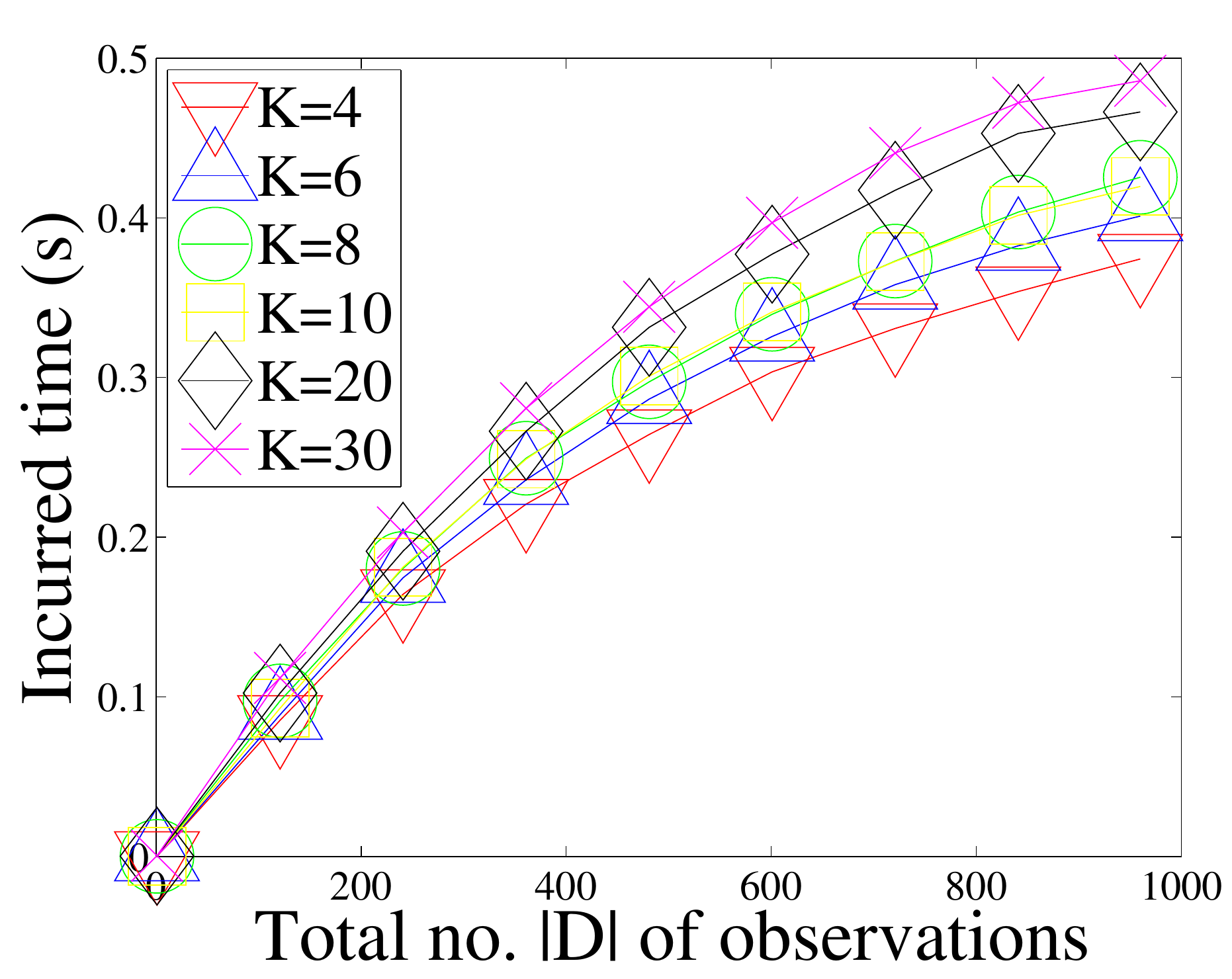}\vspace{-1mm}\\
\hspace{-2mm}(a) D$^2$FAS & \hspace{-4mm}(b) FGP &  \hspace{-4mm}(c) SoD\vspace{-3mm} 
\end{tabular}
\caption{Graphs of time efficiency vs. total no. $|D|$ of observations gathered by varying number $K$ of sensors.}
\label{fig6b}\vspace{-5mm}
\end{figure}

{\bf Scalability of data fusion.} Fig.~\ref{fig6b} shows results of the scalability of the tested data fusion methods with increasing number $K$ of sensors. In order to produce meaningful results for fair comparison, the same active sensing component has to be coupled with the data fusion methods and its incurred time kept to a minimum.
As such, we impose the use of a fully decentralized active sensing component to be performed by each mobile sensor $k$:
$w^{\ast}_k = \mathop{\arg\max}_{w_k} \mathbb{H}[Z_{Y_{w_k}}| Z_{D}]$.
For D$^2$FAS, this corresponds exactly to (\ref{dmaxent}) by setting a large enough $\varepsilon$ (in our experiments, $\varepsilon=2$) to yield $\kappa=1$; consequently, 
computational and communicational operations pertaining to the coordination graph can be omitted.

It can be seen from Fig.~\ref{fig6b}a that the time incurred by the decentralized data fusion component of D$^2$FAS decreases with increasing $K$, as explained previously.
In contrast, the time incurred by FGP and SoD increases (Fig.~\ref{fig6b}b and~\ref{fig6b}c):
As discussed above, a larger number of sensors results in a greater quantity of more informative unique observations to be gathered (i.e., fewer repeated observations), which increases the time needed for data fusion. 
When $K\geq 10$, D$^2$FAS is at least $1$ order of magnitude faster than FGP and SoD.
It can also be observed that D$^2$FAS scales better with increasing number of observations.
So, the real-time performance and scalability 
of D$^2$FAS's decentralized data fusion enable it to be used for persistent large-scale traffic modeling and prediction where a large number of observations and sensors (including static and passive ones) are expected to be available.
\begin{figure}
\begin{tabular}{ccc}
\hspace{-2.5mm}\includegraphics[scale=0.143]{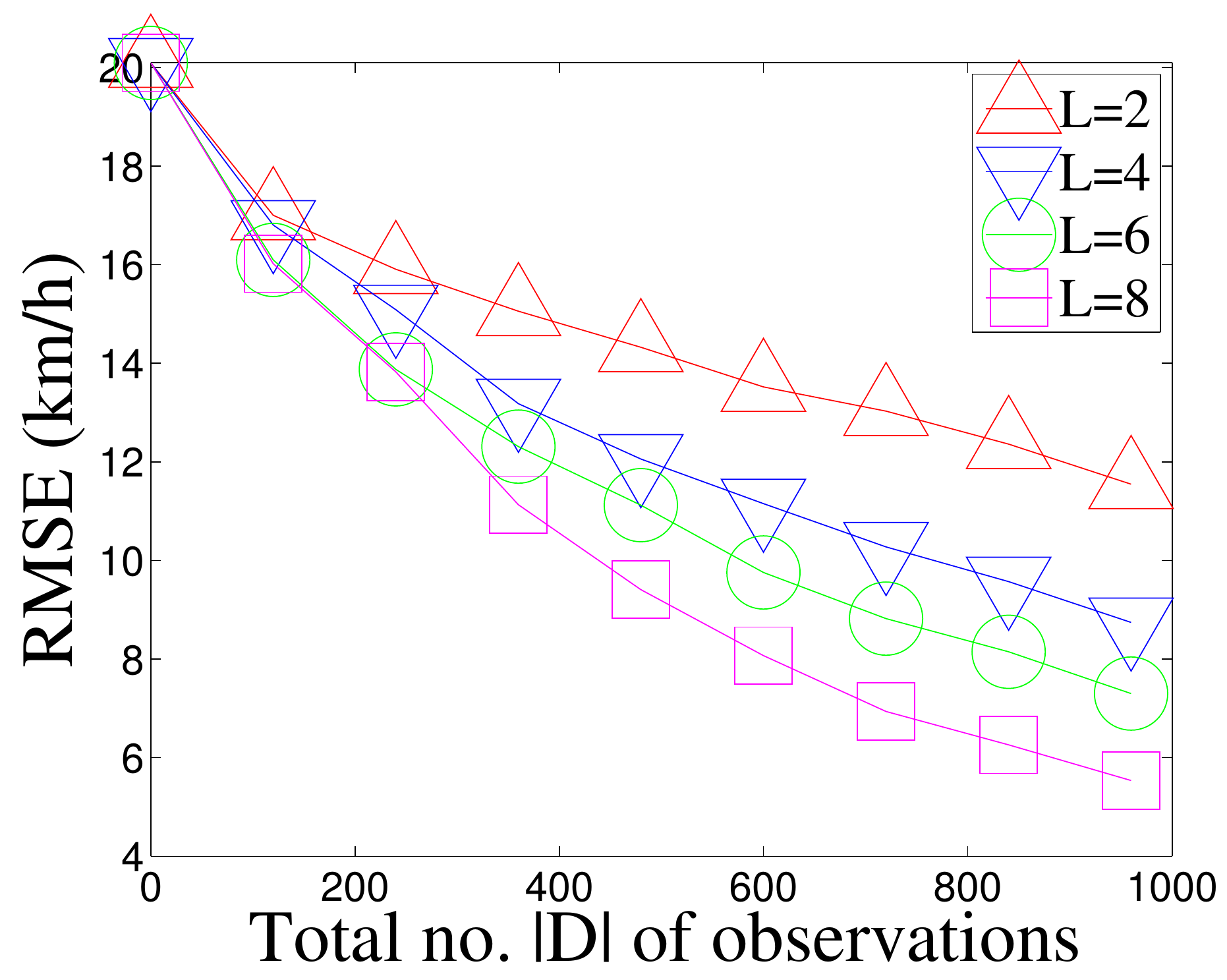} 
& \hspace{-4mm}\includegraphics[scale=0.143]{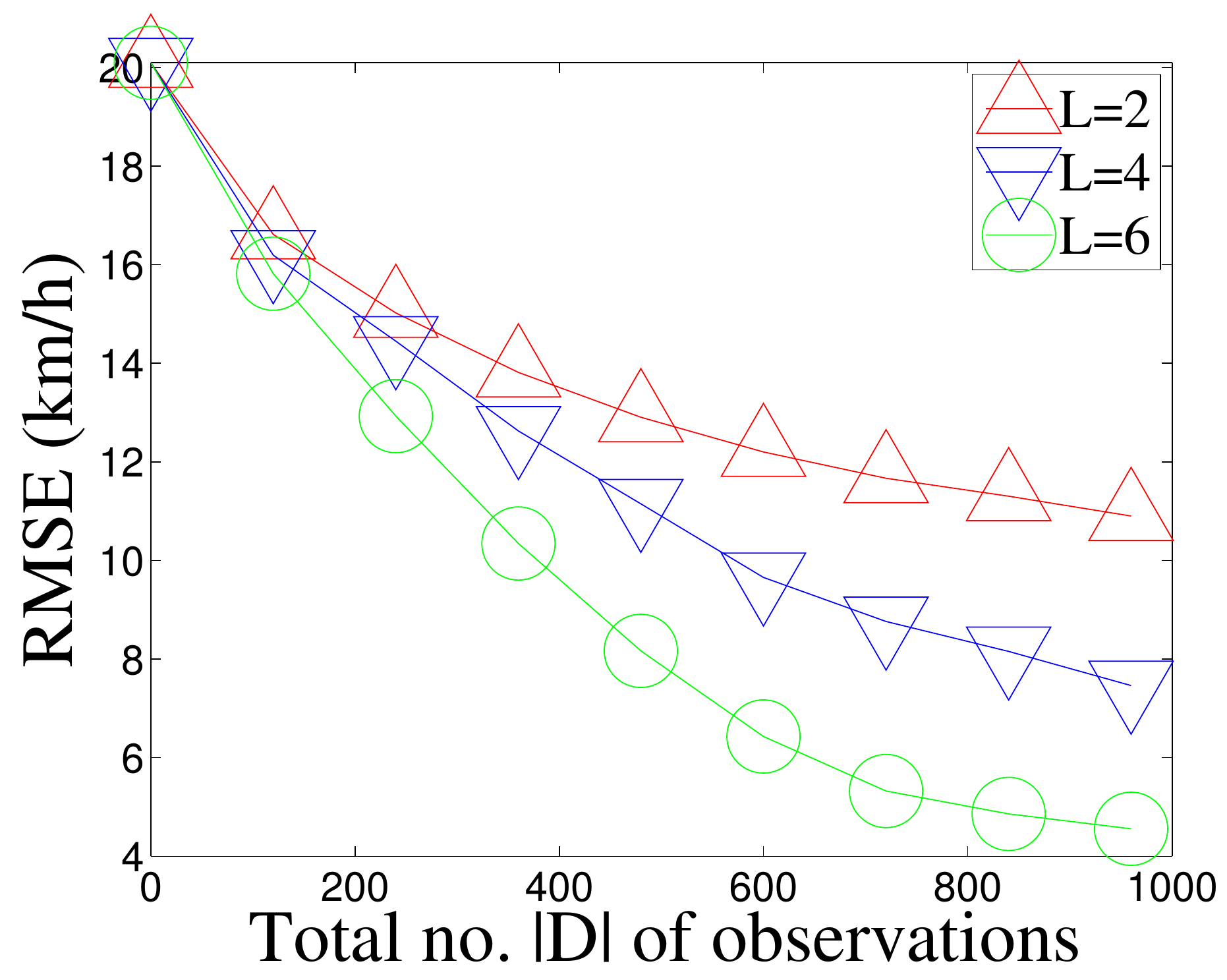} &  \hspace{-4mm}\includegraphics[scale=0.143]{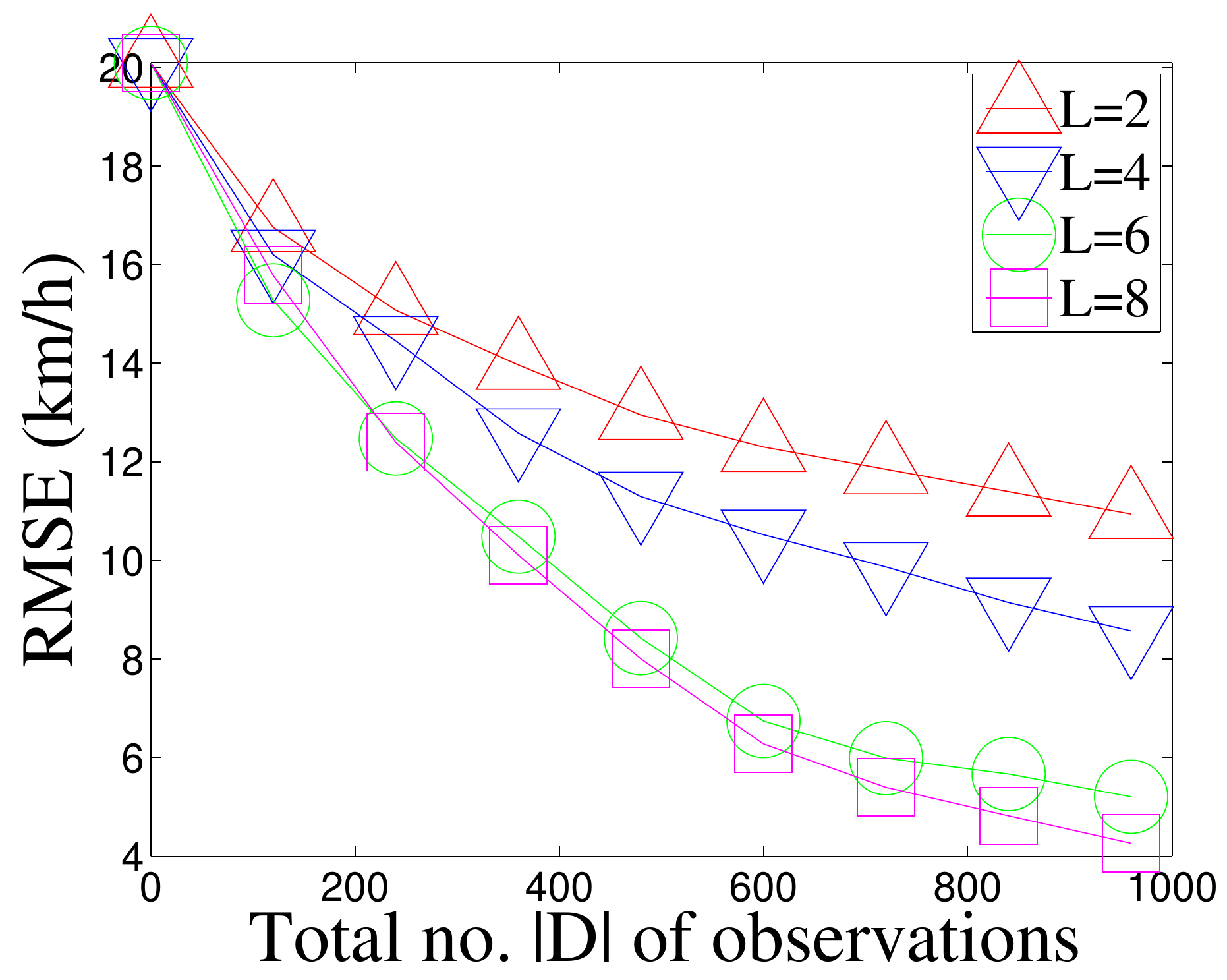}\vspace{-1mm}\\
\hspace{-2.5mm}(a) D$^2$FAS & \hspace{-4mm}(b) FGP &  \hspace{-4mm}(c) SoD\\ 
\hspace{-2.5mm}\includegraphics[scale=0.143]{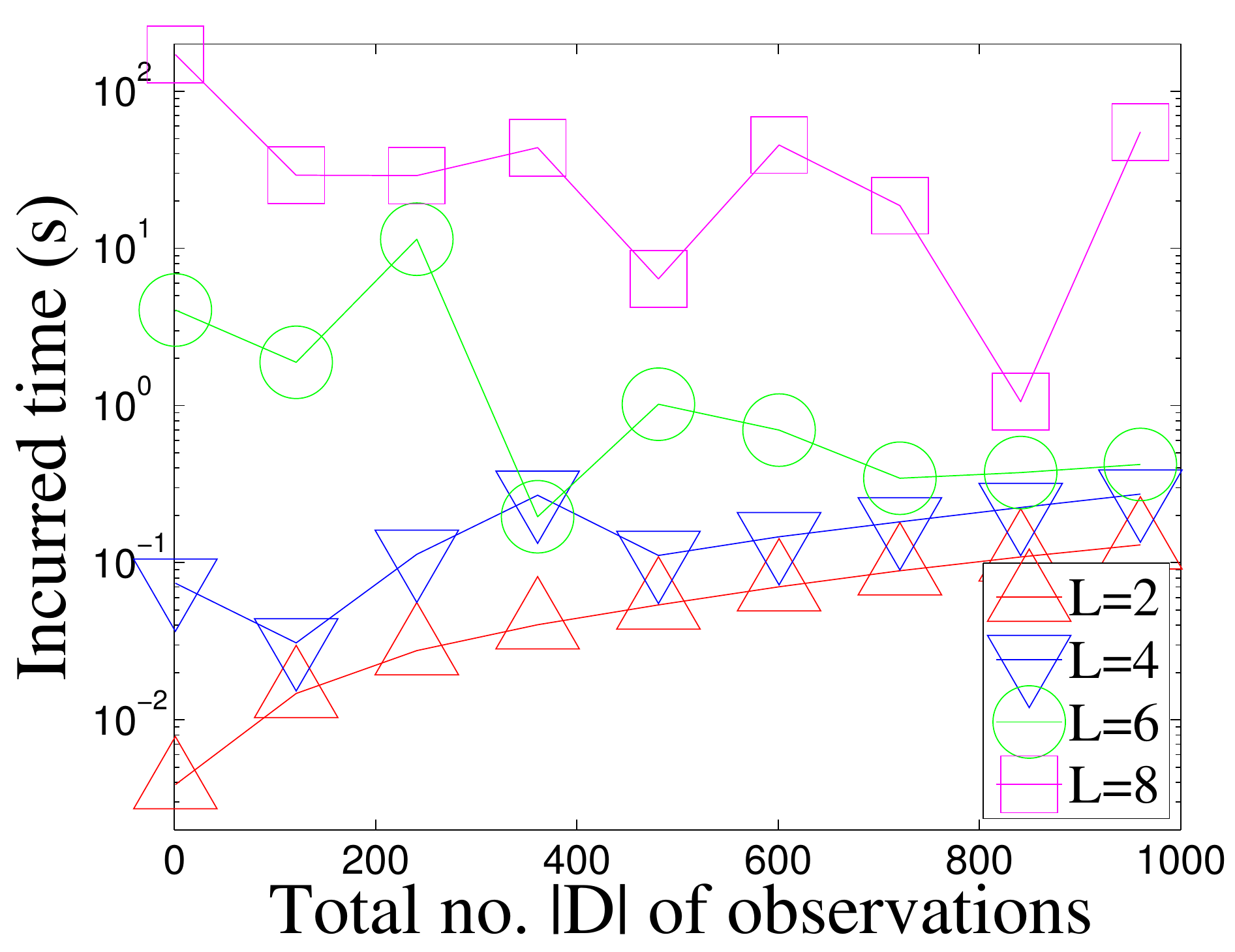} 
& \hspace{-4mm}\includegraphics[scale=0.143]{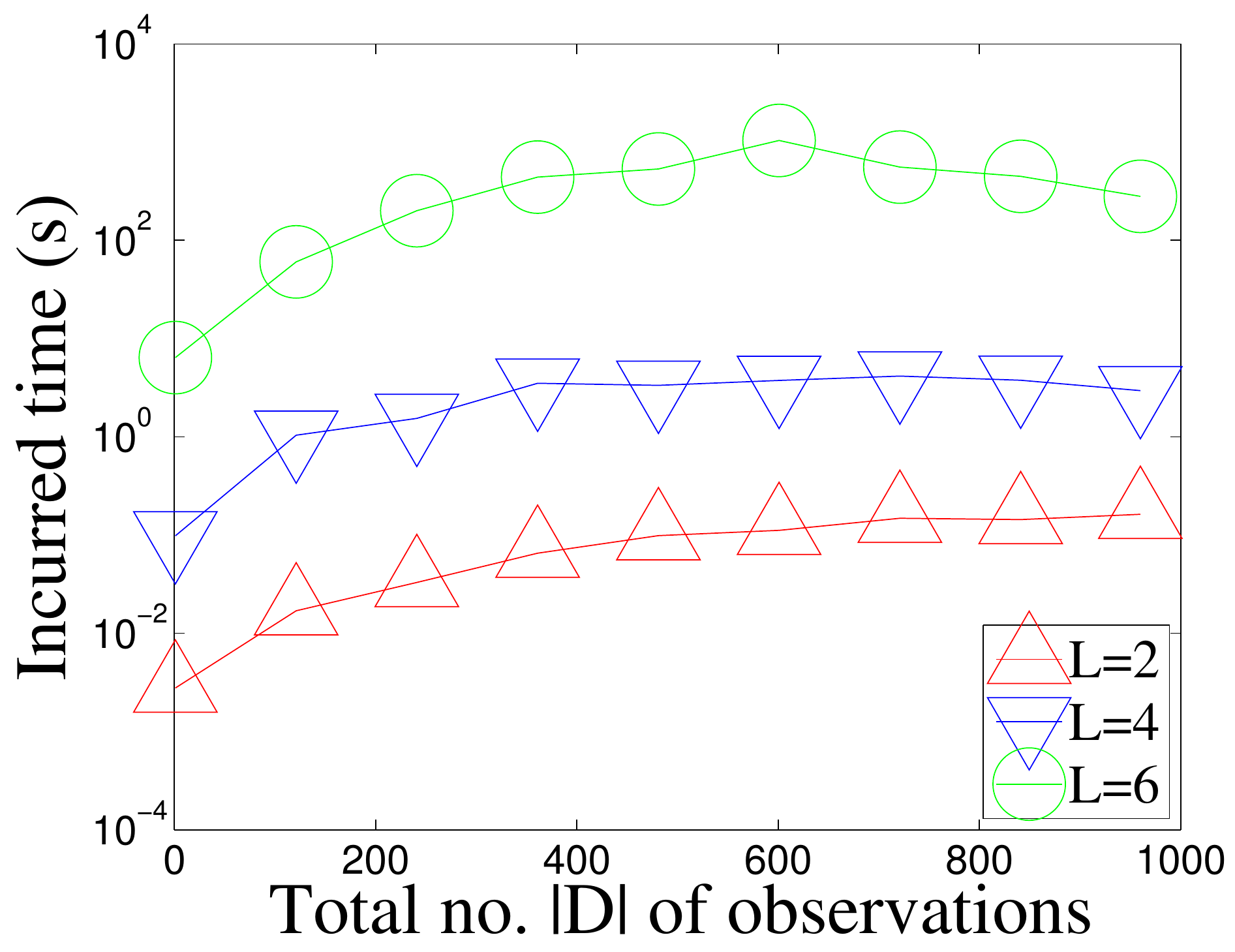} &  \hspace{-4mm}\includegraphics[scale=0.143]{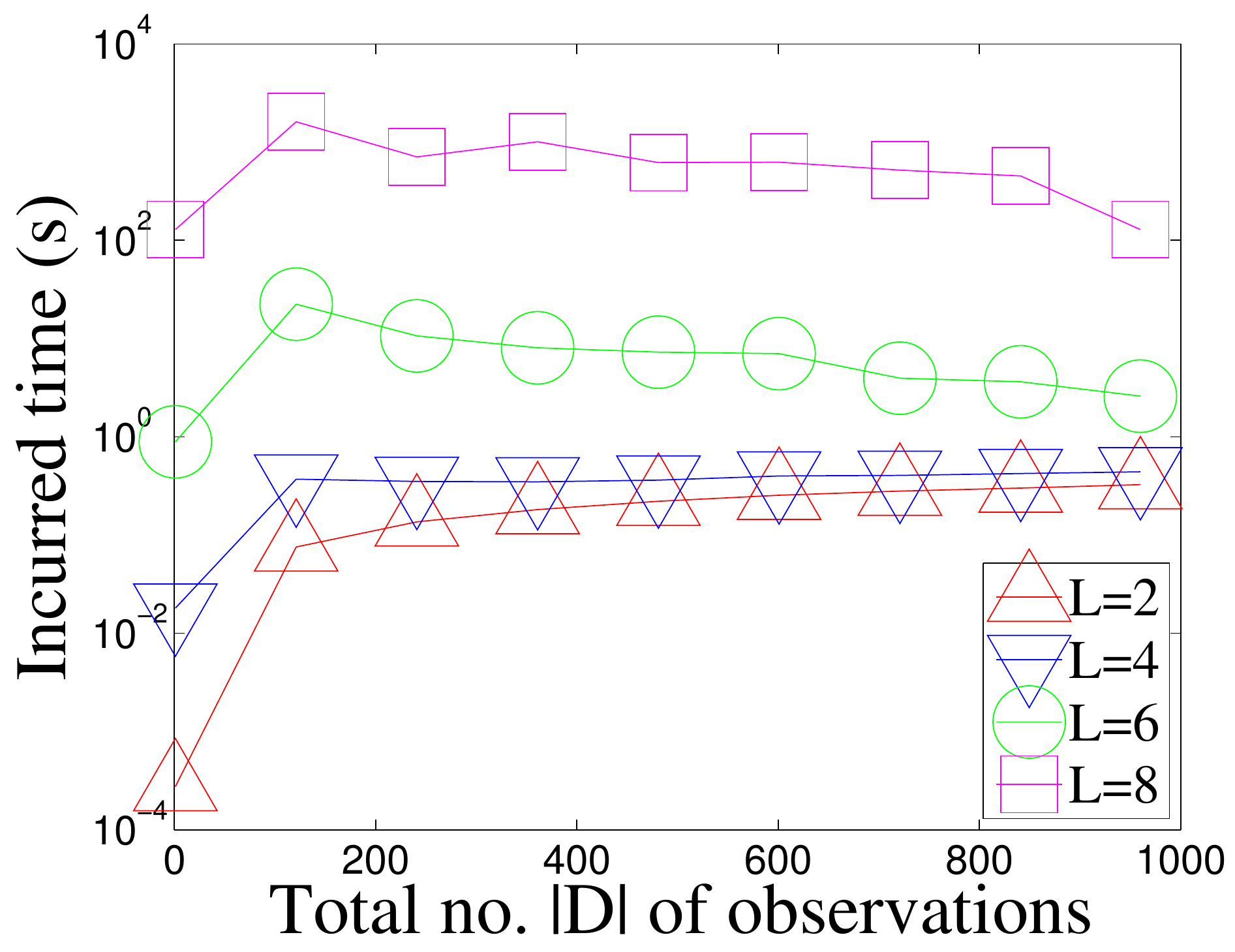}\vspace{-1mm}\\
\hspace{-2.5mm}(d) D$^2$FAS & \hspace{-4mm}(e) FGP &  \hspace{-4mm}(f) SoD\vspace{-3mm} 
\end{tabular}
\caption{Graphs of (a-c) predictive performance and (d-f) time efficiency vs. total no. $|D|$ of observations gathered by $2$ mobile sensors with varying length $L$ of maximum-entropy joint walks.}
\label{fig5}\vspace{-3mm}
\end{figure}

{\bf Varying length $L$ of walk.} Fig.~\ref{fig5} shows results of the performance of the tested algorithms with varying length $L=2, 4, 6, 8$ of maximum-entropy joint walks; we choose to experiment with just $2$ sensors since Figs.~\ref{fig6} and~\ref{fig6b} reveal that a smaller number of sensors produce poorer predictive performance and higher incurred time with large number of observations for D$^2$FAS. It can be observed that the predictive performance of all tested algorithms improve with increasing walk length $L$ because the selection of maximum-entropy joint walks is less myopic.
The time incurred by D$^2$FAS increases due to larger $\kappa$ but grows more slowly and is lower than that incurred by centralized active sensing coupled with FGP and SoD.
Specifically, when $L=8$, D$^2$FAS is at least $1$ order of magnitude faster (i.e., average of $60$~s) than centralized active sensing coupled with SoD (i.e., average of $>732$~s) and FGP (i.e., not available due to excessive incurred time).
Also, notice from Figs.~\ref{fig6}a and~\ref{fig6}d that if a large number of sensors (i.e., $K=30$) is available, D$^2$FAS can select shorter walks of $L=2$ to be significantly more time-efficient (i.e., average of $>3$ orders of magnitude faster) while achieving predictive performance comparable to that of SoD with $L=8$ and FGP with $L=6$.\vspace{-2mm}

%
\section{Conclusion}\vspace{-3mm}
This paper describes a decentralized data fusion and active sensing algorithm for modeling and predicting spatiotemporal traffic phenomena with mobile sensors.
Analytical and empirical results have shown that our D$^2$FAS algorithm is significantly more time-efficient and scales better with increasing number of observations and sensors while achieving predictive performance close to that of state-of-the-art centralized active sensing coupled with FGP and SoD.
Hence, D$^2$FAS is practical for deployment in a large-scale mobile sensor network to achieve persistent and accurate traffic modeling and prediction.
For our future work, we will assume that each sensor can only communicate locally with its neighbors (instead of assuming all-to-all communication) and develop a \emph{distributed} data fusion approach to efficient and scalable approximate GP prediction based on D$^2$FAS and consensus filters \citep{saber05a}.

{\bf Acknowledgments.} This work was supported by Singapore-MIT Alliance Research and Technology (SMART) Subaward Agreement $14$ R-$252$-$000$-$466$-$592$.

\clearpage

\bibliographystyle{natbib}
\bibliography{mybib}

\ifthenelse{\value{sol}=1}{
\clearpage

\appendix
%
%
%
\section{Proof of Theorem~\ref{thmdad}B}
\label{equ}
%
We have to first simplify the $\Gamma_{YD}\left(\Gamma_{DD}+\Lambda\right)^{-1}$ term in the expressions of $\mu_{Y|D}^{\mbox{\tiny PITC}}$ (\ref{eq:pitc.mu}) and $\Sigma_{YY|D}^{\mbox{\tiny PITC}} $ (\ref{eq:pitc.var}).
  \begin{equation} 
  \hspace{-2mm}
    \begin{array}{l} 
\left(\Gamma_{DD}+\Lambda\right)^{-1}\\
\displaystyle = \left(\Sigma_{DU}\Sigma_{UU}^{-1}\Sigma_{UD}+\Lambda\right)^{-1}\\
\displaystyle = \Lambda^{-1}-\Lambda^{-1}\Sigma_{DU}
    \left(\Sigma_{UU}+\Sigma_{UD}\Lambda^{-1}\Sigma_{DU}\right)^{-1}
    \Sigma_{UD}\Lambda^{-1}\\ 
\displaystyle =\Lambda^{-1}-\Lambda^{-1}\Sigma_{DU}\ddot{\Sigma}_{UU}^{-1}\Sigma_{UD}\Lambda^{-1} \ .
  \end{array} 
    \label{eq:pitck} 
\end{equation}
The second equality follows from matrix inversion lemma.
The last equality is due to 
\begin{equation}
    \begin{array}{l}
\displaystyle\Sigma_{UU}+\Sigma_{UD}\Lambda^{-1}\Sigma_{DU}\\
=\displaystyle \Sigma_{UU}+\sum_{k=1}^{K}\Sigma_{UD_k}\Sigma_{D_k D_k|U}^{-1}\Sigma_{D_kU}\\
= \displaystyle \Sigma_{UU}+\sum_{k=1}^{K}\dot{\Sigma}_{UU}^k =\displaystyle \ddot{\Sigma}_{UU} \ .
  \end{array} 
    \label{eq:aa} 
\end{equation}
Using (\ref{eq:sparseq}) and (\ref{eq:pitck}),
  \begin{equation} 
    \begin{array}{l} 
\Gamma_{YD}\left(\Gamma_{DD}+\Lambda\right)^{-1}\\
=\displaystyle\Sigma_{YU}\Sigma_{UU}^{-1}\Sigma_{UD}\left(\Lambda^{-1}-\Lambda^{-1}\Sigma_{DU}\ddot{\Sigma}_{UU}^{-1}\Sigma_{UD}\Lambda^{-1}\right)\\
=\displaystyle\Sigma_{YU}\Sigma_{UU}^{-1}\left(\ddot{\Sigma}_{UU}-\Sigma_{UD}\Lambda^{-1}\Sigma_{DU}\right)\ddot{\Sigma}_{UU}^{-1}\Sigma_{UD}\Lambda^{-1}\\
=\displaystyle\Sigma_{YU}\ddot{\Sigma}_{UU}^{-1}\Sigma_{UD}\Lambda^{-1}
  \end{array} 
    \label{eq:pitck2} 
\end{equation}
The third equality is due to (\ref{eq:aa}).

From (\ref{eq:pitc.mu}),
%
\begin{align*} 
 \mu_{Y|D}^{\mbox{\tiny PITC}}
&=\displaystyle\mu_Y+\Gamma_{YD}\left(\Gamma_{DD}+\Lambda\right)^{-1}\left(z_D-\mu_D \right)\\
&=\displaystyle \mu_Y+\Sigma_{YU}\ddot{\Sigma}_{UU}^{-1}\Sigma_{UD}\Lambda^{-1}\left(z_D-\mu_D \right)\\
&=\displaystyle \mu_Y+\Sigma_{YU}\ddot{\Sigma}_{UU}^{-1}\ddot{z}_U\\
&= \overline{\mu}_{Y} \ .
  \label{eq:pitc.mur}
\end{align*}
The second equality is due to (\ref{eq:pitck2}).
The third equality follows from 
$
\Sigma_{UD}\Lambda^{-1}\left(z_D-\mu_D \right)
= \sum_{k=1}^{K}\Sigma_{UD_k}\Sigma_{D_k D_k|U}^{-1}\left(z_{D_k}-\mu_{D_k} \right) = \sum_{k=1}^{K} \dot{z}^k_U =\ddot{z}_U$.

From (\ref{eq:pitc.var}),
%
\begin{equation*}
  \begin{array}{l} 
  \Sigma_{YY|D}^{\mbox{\tiny PITC}}\\
   \displaystyle=\Sigma_{YY}-\Gamma_{YD}\left(\Gamma_{DD}+\Lambda\right)^{-1}\Gamma_{DY}\\
 \displaystyle    =\Sigma_{YY}-\Sigma_{YU}\ddot{\Sigma}_{UU}^{-1}\Sigma_{UD}\Lambda^{-1}\Sigma_{DU}\Sigma_{UU}^{-1}\Sigma_{UY}\\
 \displaystyle    =\Sigma_{YY}-\Big(\Sigma_{YU}\ddot{\Sigma}_{UU}^{-1}\Sigma_{UD}\Lambda^{-1}\Sigma_{DU}\Sigma_{UU}^{-1}\Sigma_{UY}\\
 \displaystyle \ \ \ \  -\Sigma_{YU}\Sigma_{UU}^{-1}\Sigma_{UY}\Big)
    -\Sigma_{YU}\Sigma_{UU}^{-1}\Sigma_{UY}\\
  \displaystyle   =\Sigma_{YY}- \Sigma_{YU}\ddot{\Sigma}_{UU}^{-1}\left(\Sigma_{UD}\Lambda^{-1}\Sigma_{DU}
    -\ddot{\Sigma}_{UU}\right)\Sigma_{UU}^{-1}\Sigma_{UY}\\
  \displaystyle \ \ \ \   -\Sigma_{YU}\Sigma_{UU}^{-1}\Sigma_{UY}\\
    \displaystyle =\Sigma_{YY}-\left(\Sigma_{YU}\Sigma_{UU}^{-1}\Sigma_{UY}-\Sigma_{YU}\ddot{\Sigma}_{UU}^{-1}\Sigma_{UY}\right)\\
    \displaystyle =\Sigma_{YY}-\Sigma_{YU}\left(\Sigma_{UU}^{-1}-\ddot{\Sigma}_{UU}^{-1}\right)\Sigma_{UY}\\
    = \overline{\Sigma}_{YY} \ .
  \end{array} 
      \label{eq:pitc.varr} 
\end{equation*}
The second equality follows from (\ref{eq:sparseq}) and (\ref{eq:pitck2}).
The fifth equality is due to (\ref{eq:aa}).
\section{Proof of Theorem~\ref{thm2}} 
\label{equ2}
Let $\widetilde{\Sigma}_{Y_{w} Y_{w}} \triangleq \overline{\Sigma}_{Y_{w} Y_{w}} - \widehat{\Sigma}_{Y_{w} Y_{w}}$
and $\rho_w$ be the spectral radius of $\left(\widehat{\Sigma}_{Y_{w} Y_{w}}\right)^{\hspace{-1mm}-1} \widetilde{\Sigma}_{Y_{w} Y_{w}}$.
We have to first bound $\rho_w$ from above.

For any joint walk $w$, $\left(\widehat{\Sigma}_{Y_{w} Y_{w}}\right)^{\hspace{-1mm}-1} \widetilde{\Sigma}_{Y_{w} Y_{w}}$ comprises diagonal blocks of size $\left|Y_{w_{\mathcal{V}_n}}\right|\times \left|Y_{w_{\mathcal{V}_n}}\right|$ with components of value $0$ for $n=1,\ldots,\mathcal{K}$ and off-diagonal blocks of the form $\left(\overline{\Sigma}_{Y_{w_{\mathcal{V}_n}} Y_{w_{\mathcal{V}_n}}}\right)^{\hspace{-1mm}-1}\hspace{-1mm}\overline{\Sigma}_{Y_{w_{\mathcal{V}_n}} Y_{w_{\mathcal{V}_{n'}}}}$ for $n,n' =1,\ldots,\mathcal{K}$ and $n\neq n'$.
We know that any pair of sensors $k\in \mathcal{V}_n$ and $k'\in \mathcal{V}_{n'}$ reside in different connected components of coordination graph $\mathcal{G}$ and are therefore not adjacent.
So, by Definition~\ref{adj},
\begin{equation}
\displaystyle\max_{i,i'}\left|\left[\overline{\Sigma}_{Y_{w_{\mathcal{V}_n}} Y_{w_{\mathcal{V}_{n'}}}}\right]_{ii'}\right|\leq\varepsilon
\label{max2}
\end{equation}
for $n,n' =1,\ldots,\mathcal{K}$ and $n\neq n'$.
Using (\ref{max1}) and (\ref{max2}), each component in any off-diagonal block of $\left(\widehat{\Sigma}_{Y_{w} Y_{w}}\right)^{\hspace{-1mm}-1} \widetilde{\Sigma}_{Y_{w} Y_{w}}$ can be bounded as follows:
\begin{equation}
\displaystyle\max_{i,i'}\left|\left[\left(\overline{\Sigma}_{Y_{w_{\mathcal{V}_n}} Y_{w_{\mathcal{V}_n}}}\right)^{\hspace{-1mm}-1}\hspace{-1mm}\overline{\Sigma}_{Y_{w_{\mathcal{V}_n}} Y_{w_{\mathcal{V}_{n'}}}} \right]_{ii'}\right|\leq\left|Y_{w_{\mathcal{V}_n}}\right|\xi\varepsilon
\label{max3}
\end{equation}
for $n,n' =1,\ldots,\mathcal{K}$ and $n\neq n'$.
It follows from (\ref{max3}) that 
\begin{equation}
\displaystyle\max_{i,i'}\left|\left[\left(\widehat{\Sigma}_{Y_{w} Y_{w}}\right)^{\hspace{-1mm}-1} \widetilde{\Sigma}_{Y_{w} Y_{w}}\right]_{ii'}\right|\leq\max_{n}\left|Y_{w_{\mathcal{V}_n}}\right|\xi\varepsilon\leq L\kappa\xi\varepsilon\ .
\label{max4}
\end{equation}
The last inequality is due to
$\displaystyle\max_{n}\left|Y_{w_{\mathcal{V}_n}}\right|\leq L\max_{n}\left|\mathcal{V}_n\right|\leq L\kappa$. Then,
\begin{equation}
\begin{array}{rl}
\rho_w\hspace{-2mm} &\displaystyle\leq \left|\left|\left(\widehat{\Sigma}_{Y_{w} Y_{w}}\right)^{\hspace{-1mm}-1} \widetilde{\Sigma}_{Y_{w} Y_{w}}\right|\right|_2 \\
&\displaystyle\leq \left|Y_w \right| \max_{i,i'}\left|\left[\left(\widehat{\Sigma}_{Y_{w} Y_{w}}\right)^{\hspace{-1mm}-1} \widetilde{\Sigma}_{Y_{w} Y_{w}}\right]_{ii'}\right|\\
&\displaystyle\leq KL^2\kappa\xi\varepsilon\ .
\end{array}
\label{rhobd}
\end{equation}
The first two inequalities follow from standard properties of matrix norm \citep{Golub96,Stewart90}. The last inequality is due to (\ref{max4}).

The rest of this proof utilizes the following result of \cite{Ipsen03} that is revised to reflect our notations:
\begin{thm}
If 
$|Y_{w}|\rho^2_w < 1$, then
$\log\hspace{-0.5mm}\left|\overline{\Sigma}_{Y_{w} Y_{w}}\right|\leq
\log\hspace{-0.5mm}\left|\widehat{\Sigma}_{Y_{w} Y_{w}}\right| \leq \log\hspace{-0.5mm}\left|\overline{\Sigma}_{Y_{w} Y_{w}}\right|-\log\hspace{-1mm}\left(1-|Y_{w}|\rho^2_w\right)
$
for any joint walk $w$.
\label{bd1}
\end{thm} 
Using Theorem~\ref{bd1} followed by (\ref{rhobd}), 
\begin{equation}
\hspace{-6mm}
\begin{array}{rl}
\displaystyle\log\hspace{-0.5mm}\left|\widehat{\Sigma}_{Y_{w} Y_{w}}\right| - \log\hspace{-0.5mm}\left|\overline{\Sigma}_{Y_{w} Y_{w}}\right|
\hspace{-3mm} &\leq \displaystyle\log\hspace{-0.5mm}\frac{1}{1-|Y_{w}|\rho^2_w}\vspace{1mm}\\
\hspace{-3mm} &\leq \displaystyle\log\hspace{-0.5mm}\frac{1}{1\hspace{-0.5mm}-\hspace{-0.5mm}\left(K^{1.5}L^{2.5}\kappa\xi\varepsilon\right)^2}\vspace{-11mm}
\end{array}
\label{impt}
\end{equation}
for any joint walk $w$.
$$
\begin{array}{l}
\overline{\mathbb{H}}\hspace{-0.5mm}\left[Z_{Y_{w^{\ast}}}\right] - \overline{\mathbb{H}}\hspace{-0.5mm}\left[Z_{Y_{\widehat{w}}}\right]\vspace{1mm}\\
=\displaystyle\frac{1}{2}\left(
\left(\left|Y_{w^\ast}\right| - \left|Y_{\widehat{w}}\right|\right)\log\hspace{-0.5mm}\left(2\pi e\right) +
\log\hspace{-0.5mm}\left|\overline{\Sigma}_{Y_{w^\ast} Y_{w^\ast}}\right| - \log\hspace{-0.5mm}\left|\overline{\Sigma}_{Y_{\widehat{w}} Y_{\widehat{w}}}\right|\right)\vspace{1mm}\\
\leq \displaystyle\frac{1}{2}\left(
\left(\left|Y_{w^\ast}\right| - \left|Y_{\widehat{w}}\right|\right)\log\hspace{-0.5mm}\left(2\pi e\right) +
\log\hspace{-0.5mm}\left|\widehat{\Sigma}_{Y_{w^\ast} Y_{w^\ast}}\right| - \log\hspace{-0.5mm}\left|\overline{\Sigma}_{Y_{\widehat{w}} Y_{\widehat{w}}}\right|\right)\vspace{1mm}\\
\leq \displaystyle\frac{1}{2}\left(
\left(\left|Y_{\widehat{w}}\right| - \left|Y_{\widehat{w}}\right|\right)\log\hspace{-0.5mm}\left(2\pi e\right) +
\log\hspace{-0.5mm}\left|\widehat{\Sigma}_{Y_{\widehat{w}} Y_{\widehat{w}}}\right| - \log\hspace{-0.5mm}\left|\overline{\Sigma}_{Y_{\widehat{w}} Y_{\widehat{w}}}\right|\right)\vspace{1mm}\\
\leq \displaystyle\frac{1}{2}\log\frac{1}{1-\left(K^{1.5}L^{2.5}\kappa\xi\varepsilon\right)^2}\ .
\end{array}
$$
The first equality is due to (\ref{maxent21}).
The first, second, and last inequalities follow from Theorem~\ref{bd1}, (\ref{bdiaga}),
and (\ref{impt}), respectively.
}{}
\end{document}